%% file: main.tex
\documentclass[10pt,twocolumn,letterpaper]{article}

\usepackage[pagenumbers]{cvpr} 


\usepackage[accsupp]{axessibility}  

\definecolor{cvprblue}{rgb}{0.21,0.49,0.74}
\usepackage[pagebackref,breaklinks,colorlinks,allcolors=cvprblue]{hyperref}

\usepackage{multirow}
\usepackage{graphicx}
\usepackage{gensymb}
\usepackage{array}
\usepackage[dvipsnames]{xcolor}
\usepackage{tabularx} 

\def\paperID{27} 
\def\confName{EarthVision}
\def\confYear{2025}

\title{ EcoWikiRS: Learning Ecological Representation of Satellite Images from Weak Supervision with Species Observations and Wikipedia }

\author{
Valerie Zermatten$^{1*}$, 
Javiera Castillo-Navarro$^2$, 
Pallavi Jain$^{3,4,5}$,
Devis Tuia$^1$, 
Diego Marcos$^{3,5}$\vspace{0.2cm}
\and  $^1$EPFL, Sion, Switzerland 
\and  $^2$CNAM, Paris, France
\and  $^3$INRIA, Montpellier, France 
\and  $^4$CIHEAM-IAMM, Montpellier, France 
\and  $^5$Univ. of Montpellier, Montpellier, France
\and {\tt\small$^*$corresponding author: \url{valerie.zermatten@epfl.ch}} 
}

\begin{document}
\maketitle
\input{0_abstract}    
\input{1_intro}

\input{2_review}
\input{3_dataset}
\input{4_method}

\input{5_experiments}

\input{6_conclusion}

\clearpage
{
    \small
    \bibliographystyle{ieeenat_fullname}
    \bibliography{main}
}

\input{7_suppl}

\end{document}

%% file: 0_abstract.tex
\begin{abstract}
The presence of species provides key insights into the ecological properties of a location such as land cover, climatic conditions or even soil properties. We propose a method to predict such ecological properties directly from remote sensing (RS) images by aligning them with species habitat descriptions. 
We introduce the EcoWikiRS dataset, consisting of high-resolution aerial images, the corresponding geolocated species observations, and, for each species, the textual descriptions of their habitat from Wikipedia.
EcoWikiRS offers a scalable way of supervision for RS vision language models (RS-VLMs) for ecology. This is a setting with weak and noisy supervision, where, for instance, some text may describe properties that are specific only to part of the species' niche or is irrelevant to a specific image. 
We tackle this by proposing WINCEL, a weighted version of the InfoNCE loss. We evaluate our model on the task of ecosystem zero-shot classification by following the habitat definitions from the European Nature Information System (EUNIS). Our results show that our approach helps in understanding RS images in a more ecologically meaningful manner. The code and the dataset are available at \url{https://github.com/eceo-epfl/EcoWikiRS}.
\end{abstract}

%% file: 1_intro.tex
\section{Introduction}\label{sec:intro}
\begin{figure}
    \centering
    \includegraphics[width=0.8\linewidth]{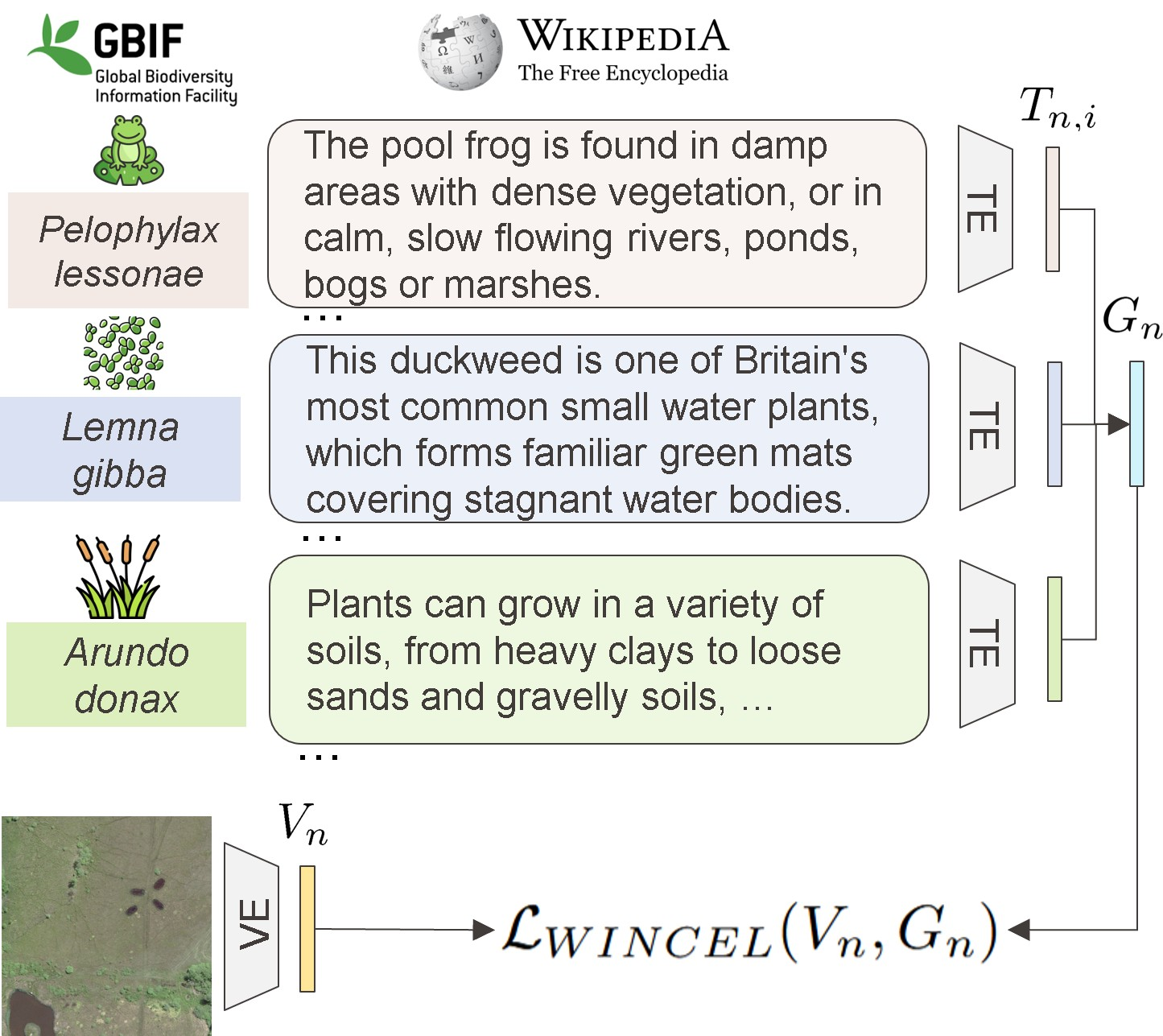}
    \caption{EcoWikiRS connects aerial images with local species observations from crowd-sourcing platforms. For each species, a habitat description from Wikipedia is retrieved and used for feature alignment. Through the proposed WINCEL loss, we learn to recognize text passages that are relevant to the images and integrating some ecological knowledge in the aerial images representation. }
    \label{fig:enter-label}
\end{figure}
Recent advances in computer vision enable large-scale analysis of satellite imagery, improving tasks like monitoring ecosystems ~\cite{lang2023high,brandt2020unexpectedly} and animal detection~\cite{delplanque2022multispecies,wu2023deep} at large scale. 
More recently, methods relying on language have appeared with the promise of making these techniques more accessible to the general audience by enabling natural language interaction with remote sensing data and thus reducing the technical barriers associated with them~\cite{chappuis2022prompt,zermatten2025learning}.  Breaking free from traditional label-based approaches, using language to interact with satellite images requires new supervision sources, to ensure the model is exposed to relevant and diverse pieces of text.  The most straightforward approaches rely on template-based~\cite{liu2024remoteclip} captions based on land cover labels or VLMs-generated text~\cite{zhang_georsclip_2024}. Other works adopted more intricate approaches based on textual labels from OpenStreetMap \cite{wang_skyscript_2024,ricci2024machine} or on the learning of an alignment from ground-level images \cite{dhakal2024sat2cap,jain_senclip_2024} without explicitly using text for training. Also, current approaches are often limited by their predominant focus on urban areas with human-centric perspectives and (land cover) object-oriented methodologies,  which can hinder the transfer of knowledge to tasks relying on a better understanding of natural or vegetal concepts.

Our study proposes to fill these gaps by learning representations for RS images integrating ecological knowledge. We achieve that by using co-located species observations from crowd-sourced portals and their description in Wikipedia as a source of supervision. The distribution of species across different regions provides key insights into the local environmental conditions they thrive in, including land cover, climatic conditions, or even soil properties. We paired each RS image with the set of observations from GBIF (Global Biodiversity Information Facility)~\cite{robertson2014gbif}, an openly available collection of geolocated species observations, including various data sources such as iNaturalist~\cite{iNaturalist}, eBird~\cite{sullivan2014ebird} or PlantNet~\cite{goeau2013plantnet}.  Using this pairing mechanism, we constructed a novel text-image dataset by associating each RS image with one or several Wikipedia articles describing the co-located species. The text describes the environmental conditions occurring at the local scale, with indications such as ``mountain and forest habitat", ``calcareous well-drained soil" or ``high altitudes and under cold conditions" that go beyond standard land cover labels (see dataset samples in Sup.~\ref{tab:dataset_samples}). 
Learning from such a dataset presents several challenges. First, the presence of noisy pairs is unavoidable, either due to species misidentification or incorrectly geolocated observations. Moreover, some species are considered generalist species and adapt to many different habitats (e.g. the house sparrow), and might be observed in multiple habitats. This means that all locations where a house sparrow has been observed will be associated with the same set of sentences, resulting in a weak and noisy learning signal. Noisy image-text pairs also occur in general VLMs datasets, although image captions generally describe specific aspects of the image. In such cases, either the dataset size is considered sufficient to ignore the noise effect~\cite{radford2021learning,jia2021ALIGN}, or the noise can be explicitly filtered out with a specific approach~\cite{li2022blip,wei2023instructiongpt}.

Here, we explore these challenges for a dataset marrying ecological observations with remote sensing images and Wikipedia text. We create a triplets dataset (image, GBIF observations, Wikipedia texts) over Switzerland and study methods for learning alignment under weak supervision, including a loss function to best learn domain-specific features from the supervision of weak image-text pairs.  Unlike previous work connecting Wikipedia with satellite images \cite{daroya_wildsat_2024,uzkent_learning_2019},  we work with very high-resolution imagery and carefully select text that is ecosystem-related  and relevant to the given image. We evaluate our approach in a zero-shot setting for the task of ecosystem mapping, a task requiring domain-specific knowledge. We also show qualitative examples of our model capacities by mapping the presence of ecological concepts across Swiss landscapes.

%% file: 2_review.tex
\begin{figure*}[ht!]
    \centering
    \includegraphics[width=0.9\linewidth]{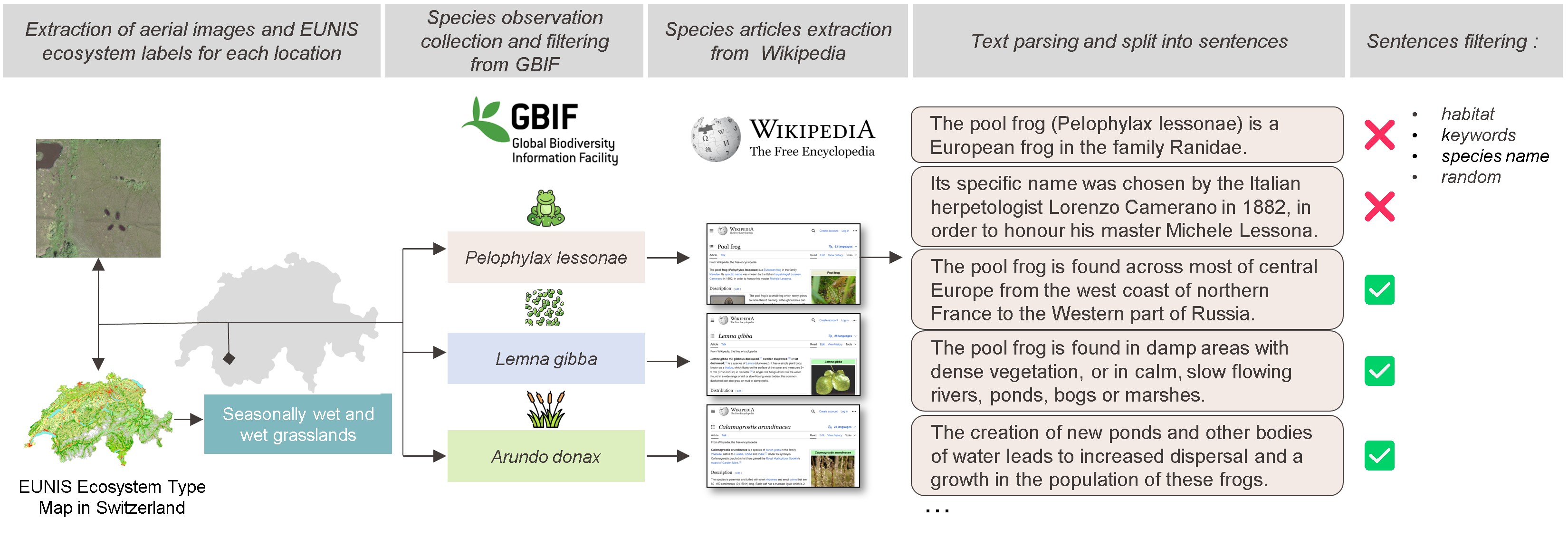}
    \caption{EcoWikiRS dataset preparation. For each location containing species observation in GBIF, an aerial image and its ecosystem type from the EUNIS map are extracted. The corresponding Wikipedia articles are collected. The retrieved text is parsed and split into sentences before a filtering step aiming at keeping only ecosystem-related sentences. }
    \label{fig:dataset_fig}
\end{figure*}
\section{Related Works}
\label{sec:review}
\noindent\textbf{Vision-Language Foundation Models for remote sensing.}
VLMs trained with Web scale image-text data, such as  CLIP~\cite{radford2021learning} and ALIGN~\cite{jia2021ALIGN}, break free from the traditional approach using a fixed set of categories in visual recognition tasks and learn visual features from natural language supervision. 
Direct application of these models to RS is on the rise, but remains challenging due to the domain shift from natural images to overhead imagery. For instance, RemoteCLIP~\cite{liu2024remoteclip} focused on existing RS datasets, and is thus limited by the scale and semantic diversity of the existing text labels.
Other models tackled this issue by generating diverse captions with Large Language Models (LLMs)~\cite{chu_towards_2024_geotext}, relying on existing image-text pairs of satellite images~\cite{zhang_georsclip_2024}, exploiting geolocation to bind RS images with other sources of supervision such as OpenStreetMap land cover labels~\cite{wang_skyscript_2024,ricci2024machine} or ground-level imagery~\cite{jain_senclip_2024,dhakal2024sat2cap}. 
Unlike these works, we focus on introducing knowledge about the ecological value of a place in the VLMs, going beyond standard land cover mapping concepts.

\noindent\textbf{Learning Vision-Language alignment from weak supervision.} 
In early VL models~\cite{radford2021learning,jia2021ALIGN}, the  Web scale of the training datasets is often considered sufficient to compensate for the weak alignment of cross-modal pairs. A  contrastive loss, such as the InfoNCE~\cite{oord2018representation} is used as supervision to bring positive pairs together and push negative pairs apart. To improve performance and data efficiency, later works complement this objective with self-supervision~\cite{mu_slip_2021,li_supervision_2022_declip}, add cross-modal losses such as image-text matching~\cite{li2023blip2},  or use nearest neighbor supervision~\cite{el2023language_guided,dwibedi_little_2021}. 
To explicitly take into account the problem of learning from noisy text-image pairs, some works propose to relax the ``hard" one-hot encoded targets of the positive pairs in the contrastive loss by attributing some probability to negative pairs.  
Label smoothing encourages the model to be less confident in the hard targets by adding a small but uniform probability to all classes. Bootstrapping~\cite{reed2014training} integrates the model posterior distribution in the targets, with either the logits or the predictions. The knowledge from pretrained image and sentence encoders can be used to produce pseudo labels~\cite{cheng2021data,gao_softclip_2024}, filter the text paired with the images~\cite{li2022blip,yang_alip_2023} or select the most confident samples~\cite{li2022selective}. 
To reduce the impact of outliers, RINCE~\cite{morgado2021robust} rescales the sample importance based on estimated noise level.

Compared to the general domain, the filtering of noisy text-images pairs in RS-VLMs has been little explored outside of dataset creation.  Several works~\cite{zhang_georsclip_2024,wang_skyscript_2024} filter image-text pairs with CLIP~\cite{radford2021learning} to remove text that is not sufficiently relevant to the image. Other works focus on human verification through direct visual verifications or with human-computer  annotation systems ~\cite{chu_towards_2024_geotext,shabbir2025geopixel,muhtar2024lhrs}. 
Here, we introduce WINCEL, a weighted InfoNCE loss function that relies on a pretrained models' knowledge, to explicitly filter out noisy text-images pairs through assigning a weight to each pair that depends on their current alignment strength (see Section~\ref{sec:method}).

\noindent\textbf{Learning from species descriptions and Wikipedia.}
Species observations data based on citizen science are widely used for conservation research, such as for species distribution models~\cite{van2024regional,callaghan2023unveiling} or for plant functional traits mapping~\cite{wolf2022citizen}. Despite the presence of numerous biases and incorrectly labeled observations~\cite{zizka2020no}, they allow mapping at large scale in a timely manner with reasonable costs~\cite{kirchhoff2021rapidly,vardi2021inaturalist}.
Species observations have been combined with textual descriptions to improve their recognition performances.
The ZEST~\cite{paz-argaman_zest_2020} dataset combined bird pictures with their Wikipedia descriptions and demonstrated a better capacity to identify bird species in a zero-shot manner. \cite{saha_improved_2024} improved the zero-shot classification of different species by generating fine-grained descriptions of their appearance and characteristics with LLMs. 

Studies using the iNaturalist dataset~\cite{iNaturalist} have shown the benefits of using species observation metadata such as geolocation or aerial images for improving species identification~\cite{berg2014birdsnap,mac2019presence}. In this work, we reciprocally use species observations to gain insights into their living environment. For species range maps estimation, LE-SNIR~\cite{hamilton_2024_combining}  proposes a model predicting presence based on a single data point, thanks to textual description of the species habitat and range.  TaxaBind~\cite{sastry_taxabind_2024} unifies six ecological data types, including RS images, taxonomic names and ground-level images, and improves the classification and cross-modal retrieval performances on various geo-aware ecological tasks.  
Closer to our work, WildSAT~\cite{daroya_wildsat_2024} learns satellite image representations by aligning them with descriptions of species observed by iNaturalist~\cite{iNaturalist} users. Their work combines satellite images at different time steps with a random section of the Wikipedia article, geolocation and climate information. They improve RS foundation model performances on land cover mapping tasks. Unlike their work, we utilize very high-resolution aerial images and thus aim at a much more fine-grained text-image alignment. Moreover, we focus on extracting ecologically meaningful sentences from Wikipedia and evaluate our approach on tasks relative to the natural concepts. 

%% file: 3_dataset.tex
\section{EcoWikiRS Dataset}
\label{sec:dataset}
In this section, we present the EcoWikiRS dataset and detail the steps of its creation and refinement (Figure~\ref{fig:dataset_fig}).

We collected a dataset over Switzerland composed of very high-resolution aerial images (swissIMAGE product, $50 cm$ resolution), species observations from GBIF~\cite{robertson2014gbif} and sentences from Wikipedia describing the observed species. The species observations are paired with the images by location, i.e. their geolocation is within the area covered by the aerial image.  Wikipedia sentences describing all the observed species are extracted, filtered and assigned to the image, forming a dataset composed of triplets:
each aerial image $I_n$ is associated with a list of observed species, and a set $J_n = \{s_{n,k}\}_{k=0}^K$ of $K$ Wikipedia sentences describing the observed species. 
The EcoWikiRS dataset contains a total of $N=91'801$ aerial images associated with one or several species out of $2'745$ different species. The dataset is split into train ($60\%$), validation ($10\%$) and test ($30\%$) following a spatial block split with blocks of $20$ km to avoid spatial autocorrelation (See Suppl. Figure~\ref{fig:block_split}). 
In the following, we describe each data source.  Dataset samples and additional statistics are given in the Supplementary material.

\subsection{Data sources}\label{sec:sources}
\noindent\textbf{Species observation from GBIF.}
Geolocated species observations are collected throughout Switzerland from the Global Biodiversity Information Facility (GBIF) Web portal~\cite{robertson2014gbif}. GBIF  gathers observations from crowd-sourcing platforms, such as iNaturalist~\cite{iNaturalist}, governmental agencies or scientific surveys. Species observations undergo a filtering procedure (see Suppl. Section~\ref{app:gbif_filtering}). We selected plants and animal species and discarded bacteria, algae or fossil specimens; we removed species with geo-location uncertainty above 100 m, incomplete metadata (e.g. missing species identification) or species without Wikipedia articles. Following this filtering stage, $274'241$ observations were selected among the $18$ million occurrences downloaded~\cite{gbif_download}.  

\noindent\textbf{Aerial images.}
We used the openly available swissIMAGE~\footnote{\url{https://www.swisstopo.admin.ch/fr/orthophotos-swissimage-10-cm}} product of the Federal Office of Topography. Aerial images were acquired in summer in the years $2020-2022$.  RGB bands are available at a spatial resolution of 10 cm, which we down-sample to 50 cm. Images were further split into tiles of 100 m by 100 m, following the grid of  EUNIS ecosystem type map~\cite{chytry2020eunis}. We retained aerial images with at least one observation from GBIF within its footprint.

\noindent\textbf{Wikipedia descriptions.}
Textual descriptions of the species are retrieved based on the species binomial name from an English Wikipedia dump~\cite{wikipedia2025}. For each species, the article is cut into sections and irrelevant sections are removed (e.g. ``See also", ``Gallery" or ``Bibliography"). The resulting text is further split into individual sentences based on a set of parsing rules. Different sets of sentences are extracted and will be compared in the experiments (e.g. Table~\ref{tab:Table2}): (1) \emph{habitat}: sections whose title contains words related to ecological preferences such as ``habitat", ``distribution", ``cultivation", ``ecology" or ``range",  (2) \emph{keywords}: sentences containing at least one keyword from a predefined list of ecology related concepts such as ``wet", ``alpine", ``calcareous", etc. (see list of keywords in the Suppl. Section~\ref{sec:keywords}), (3) \textit{species binomial name} such as ``Lemna gibba", (4) \emph{random}: all sentences.

\subsection{EUNIS habitat maps as downstream task.}
To validate the zero-shot classification capabilities of our model, we use the EUNIS habitat classification framework~\cite{chytry2020eunis}, a hierarchical European system for habitat identification based on plant species communities.  Its classes cover all types of habitats found in Europe.  We use the \textit{Ecosystem Type Map v3.1}~\cite{weiss2018ecosystem} layer with a spatial resolution of 100 m and the EUNIS classes at Level 2. We merged or removed habitats with less than 100 occurrences and we under-sampled habitats with more than 10k occurrences to balance the distribution of classes. Wetlands ecosystems were merged at level L1 due to their rare occurrence. A final set of 25  habitats is present in Switzerland (see Suppl. Figure~\ref{fig:eunis_habitatl} ) and the distribution of samples into habitats is shown in the Suppl. Figure~\ref{fig:barplot_distr_eunis}.

%% file: 4_method.tex
\section{Method} 
\label{sec:method}
In this work, we propose to use the rich and text-aligned image representations provided by pretrained RS-VLMs and enrich them with ecological knowledge from the EcoWikiRS dataset. The image encoder is updated with the WINCEL loss, a contrastive approach we propose for learning domain-specific concepts from noisy/weak text supervision. 
We start with the hypothesis that, although several sentences in a Wikipedia article will be unrelated to the image they are paired with, at least one sentence per article is likely to describe the content of the image.  
We develop a loss function that aims at recognizing and learning from sentences related to the visual features present in the image, while ignoring other sentences, similarly to~\cite{paz-argaman_zest_2020}. Inspired by RINCE~\cite{morgado2021robust}, we design the WINCEL loss by  accounting for two types of noise in text-image pairs: 
\newline
\noindent 1) A \textbf{false positive pair} occurs when the signals from different modalities are uninformative of each other. This type of noise is frequent in the EcoWikiRS dataset since the sentences are assigned to the image depending on the presence of species rather than on the visual content of the image. Especially for generalist species, which can live in several habitats, the potential of appending uninformative sentences can be high. Therefore, we expect the false positive pairs to outnumber the true positive ones. 
\newline
\noindent 2) Within a batch, the text associated with the other images is used as the negative samples, but such text will often include \textbf{false negatives}, i.e. texts that are appropriate to describe the image, even though they are not paired with it.   This problem is also exacerbated in EcoWikiRS, where two images with observations of the same species will be associated to the same set of sentences.

\subsection{Weighted InfoNCE loss (WINCEL)}
We address these issues by learning a weighting function that estimates, for each image $I_n$, the relevance of each sentence $s$ from $J_n$. This approach is designed to directly down-weight false positive pairs by relying on the model's current knowledge and focusing on the most confident image-sentence pairs, similarly to~\cite{li2022selective}. This also brings the text representation closer to the image, thus relieving the issue of false negatives as well.
We use a visual encoder $f_v$ and a text encoder $f_t$ including unit-normalization, to obtain the respective embeddings $V_n=f_v(I_n)$ and $T_{n,k}=f_t(s_{n,k})$. 
The InfoNCE loss \cite{oord2018representation,radford2021learning} is a standard loss function to optimize the alignment of a visual embedding $V_n$ with a unique corresponding textual embedding $T_n$ that aims at reducing their cosine distance compared to other embeddings. 
The alignment is controlled by the temperature hyperparameter $\tau$, where a low value makes the posterior distribution peakier and more specific to the given embedding in the pair assignment. For a random sentence $s_{n,k}$ paired with image $I_n$, we have:
\begin{equation} \label{eq:infoNCE}
 \mathcal{L}_{con} (V_n,T_n) = - \log \frac{ \exp{  ( V_n \cdot T_n / \tau} )}{ \sum_{j=1}^{N} \exp{( V_n \cdot T_j / \tau)} }
\end{equation}
Although this formulation, which encourages relative alignment only, is designed to deal with noisy pairs, our setting includes an extra source of noise since the text is assigned to each image without accounting for its visual content. 
Instead of learning from a single sentence paired with image $I_n$, we build a weighted text representation $G_n$ that depends on several sentences.  $G_n $ is built as a linear combination of the sentence embeddings $T_{n,k}$ associated with image  $I_n$: 
\begin{equation} 
G_n = \sum_{k=0}^K  \alpha_{n,k} \cdot T_{n,k} \label{eq:G1} 
\end{equation}
where $\alpha_{n,k}$ is a scalar weight accounting for the importance of the sentence embedding $T_{n,k}$ in matching the content of the image.  
We defined $\alpha_{n,k}$ based on the cross-modal similarity between the image and each sentence embedding, rescaled with a softmax function $\sigma$:
\begin{equation} 
\alpha_{n,k} = \sigma( V_n \cdot T_{n,k} / \tau) \label{eq:a}
\end{equation}
Combining~\eqref{eq:G1} and~\eqref{eq:a}, the weighted text representation $G_n$ becomes:
\begin{equation} \label{eq:G_n}
 G_n =  \sum_{i=0}^K  \sigma( V_n \cdot T_{n,k} / \tau) \cdot T_{n,k}  
\end{equation}
This formulation relies on the pretrained model's knowledge to learn better representations. Moreover, it can be seen as a form of text augmentation similar to mixup~\cite{zhang2018mixup}, where a linear interpolation is operated in the textual space, but here based on more than 2 embeddings. The proposed WINCEL is : 
\begin{equation} \label{eq:wincel}
 \mathcal{L}_\text{WINCEL} (V_n,G_n) = - \log \frac{ \exp{  ( V_n \cdot G_n / \tau} )}{ \sum_{j=1}^{N} \exp{( V_n \cdot G_j / \tau)} }
\end{equation}

%% file: 5_experiments.tex
\section{Experimental setup}
\label{sec:experiments}

\noindent\textbf{Training procedure and implementation details.}
We fine-tune four different VLM backbones, namely RemoteCLIP~\cite{liu2024remoteclip}, SkyCLIP~\cite{wang_skyscript_2024}, GeoRSCLIP~\cite{zhang_georsclip_2024} and CLIP~\cite{radford2021learning}, starting with ViT-B/32 pretrained backbones available on their respective online repositories for the RS-VLMs, and from the \textit{openclip}~\cite{ilharco_gabriel_openclip} library trained on the \textit{LAION-2B} dataset~\cite{schuhmann2022laion}. We compare the performance of the pretrained models with results after fine-tuning with a standard InfoNCE loss and with our proposed WINCEL using the EcoWikiRS dataset. Unless stated otherwise, we use the ``habitat" sentences described in Section~\ref{sec:sources}. 

During model training, we observed that fine-tuning the positional encoding layer and the projection head of the visual encoder in ViT-B/32, while keeping the text encoder frozen, led to the best performances (see Supp. Table~\ref{tab:table6}). The AdamW~\cite{loshchilov2017ADAMWdecoupled} optimizer is used with an initial learning rate of $1e^{-4}$. The training is carried out for 60 epochs, with a batch size of $256$ and a step scheduler with a step size of $2$ and a decay multiplier of $0.95$. We set the number of sentences per samples to $K=15$, and use padding when fewer than $K$ sentences are available, i.e. text features are set to zero. The temperature hyperparameters are set to $0.07$ for the InfoNCE loss and $0.15$ for WINCEL after a grid search performed separately for each method. Center cropping, random flip, rotation and color augmentation were used on all approaches.

\noindent\textbf{Zero-shot ecosystem prediction.} To evaluate if our approach learns relevant ecological knowledge,  we study the performance of WINCEL and EcoWikiRS on the task of zero-shot ecosystem type prediction. 
Given an aerial image from the test split as input, the text encoder is prompted directly with the $25$ labels of the EUNIS categories. The class with the highest text-image cosine similarity is chosen. We observed that the use of specific templates or longer class descriptions as prompts does not lead to better performances, as shown in the Supp. Table~\ref{tab:Table5}. We measure performance with overall accuracy (OA) and macro-averaged F1-score (F1), which evaluate the model's overall performance and the mean performance per class, respectively. The results of a supervised upper-bound based on the SkyCLIP image encoder and fine-tuned with a cross-entropy loss are also shown as a comparison.
\section{Results and discussion}
In this section, we present quantitative and qualitative results on EcoWikiRS with the proposed WINCEL. Quantitative results include zero-shot predictions of the EUNIS ecosystem labels on the EcoWikiRS dataset, as well as an ablation study of the proposed training strategy and of the type of text used as input. Qualitative results are cross-modal retrieval examples. 
\subsection{Zero-shot classification on EUNIS}
Table~\ref{tab:table1} summarizes the zero-shot classification performance for different pretrained models, the InfoNCE loss and the proposed WINCEL loss. 
Fine-tuning on the EcoWikiRS dataset consistently outperforms the pretrained models,  demonstrating that our dataset allows models to learn features relevant for this task.  
The proposed WINCEL approach is better than InfoNCE for three out of four VLMs, illustrating its capacity to focus on more useful sentences during training. While SkyCLIP, CLIP and GeoRSCLIP obtain similar levels of performance after fine-tuning, RemoteCLIP performs poorly both in pretrained, at $11.9\%$ overall accuracy, and fine-tuned approaches, at $20.9\%$ with WINCEL. This is possibly due to a semantic gap, as RemoteCLIP training set has a strong focus on artificial land cover, and its captions are shorter and very different from Wikipedia sentences. The fact that fine-tuning with WINCEL results in lower performance than InfoNCE when initialized with RemoteCLIP could then be due to the fact that the sentence encoder is not able to align the representations to those of ecosystem types. \\
On the contrary, CLIP which has not been specifically trained on RS imagery but was exposed to a large breadth of semantic concepts, is able to profit substantially from fine-tuning, increasing from $14.7\%$ OA to $30.9\%$ when using WINCEL. With the standard InfoNCE, the OA only increases to $25.3\%$.
Interestingly, the performance of SkyCLIP, whose training set includes swissIMAGE imagery (i.e. our source of aerial imagery), does not stand out compared to the model fine-tuned on other datasets, again suggesting that the gap is more related to semantics than to sensor characteristics.
The best-performing pretrained model is GeoRSCLIP, although fine-tuning it, regardless of whether with InfoNCE or WINCEL, results in lower performances than fine-tuning CLIP or SkyCLIP.
This, along with the fact that pretrained SkyCLIP performs better than CLIP, led us to use SkyCLIP for the next studies presented in this paper. The comparison with the supervised baseline highlights that further improvements are still possible, as the EcoWikiRS dataset offers only weak supervision. 
\subsection{Ablation studies}
\input{table1}
\input{table2} 
\textbf{Type of input text.}
We trained using different sets of text from Wikipedia articles, including species names, sentences from the habitat section, sentences containing certain keywords and random sentences (as described in  Section~\ref{sec:dataset}). The results in Table~\ref{tab:Table2} show that passages from the ``habitat" section consistently outperform the other approaches. The species name obtains the worst results, likely due to its limited capacity to convey sufficient contextual or descriptive information relevant to the image. This suggests that the targeted selection of relevant content can enhance the effectiveness of the representations learned, compared to the full article.  The ``random" sentences contain almost seven times more sentences than the ``habitat" texts (See suppl. Table~\ref{tab:sentences_count}),  highlighting the importance of quality over quantity for improving model performance. 
\input{table4}
\input{fig_prompts}

\noindent\textbf{Fine-tuning approach.} We combine several approaches for handling noisy labels with our proposed WINCEL loss and compare them to a standard approach using the InfoNCE loss. Two groups of alternative strategies are considered. We implement the bootstrapping strategy for noisy labels following~\cite{reed2014training}, and fine-tune the models using a weighted combination of the original one-hot encoded targets and the model predicted logits (\textbf{bootstrap-soft}) or targets (\textbf{bootstrap-hard}). 
The weights for the linear combination of targets were empirically found and set to 0.8 for bootstrap-soft and 0.9 for bootstrap-hard.
Sampling strategies involve picking the sentences with the highest cross-modal similarity (\textbf{top-1 sampling}), while \textbf{top-p} sampling implies choosing one sentence by using the cross-modal similarity as the probability of being chosen.  \textbf{Sub-string augmentation} consists of randomly sampling a subset of consecutive words from a sentence, i.e., picking a sub-sentence of $3$ to $15$ words from the original sentence as a form of text augmentation.
\\
Results for SkyCLIP trained with the ``habitat'' texts are shown in Table~\ref{tab:table4}. Regarding sampling techniques, the top-1 sentences sampling shows improvements in both OA and F1-scores compared to the InfoNCE approach. Top-p sentences sampling and sub-string augmentation show a clear decline in metrics, likely because they introduce too much noise during the learning process. Bootstrap-hard performs better than bootstrap-soft in terms of OA and F1 score, while still under-performing compared to our proposed WINCEL. These results demonstrate that naive InfoNCE applied to random text is suboptimal. Instead, filtering text prior to contrastive learning yields superior performance which emphasizes the importance of high-quality text selection for the cross-modal alignment.
\subsection{Visual results}
\input{table_simi}

To illustrate some ecological concepts learned by our model, we geographically visualize interactions between the RS images and sentences describing habitats. We generate visual features with both the pretrained and the fine-tuned SkyCLIP model and plot the cross-modal similarity on the surface of Switzerland (one image of 100 m by 100 m per km$^2$). Figure~\ref{fig:prompts} shows the maps for the pretrained (top) and fine-tuned (bottom) models for six Wikipedia sentences.  Figure~\ref{fig:prompts} (a) and (e)  serve as references for temperature patterns and general land cover in Switzerland.
Overall, we observe that the maps generated by the fine-tuned models are more coherent than those of the pretrained model, which we assess by using the reference maps as proxies.  In Figure~\ref{fig:prompts} (b),  the fine-tuned model (bottom) correctly highlights the warmest region of Switzerland, while the pretrained model map (top) does not show a significant trend. Similarly, Figure~\ref{fig:prompts} (c) shows no discernible pattern for the pretrained model (top), while the fine-tuned model (bottom) correctly highlights the Swiss plateau (low lands). For plots (d), (g) and (h), both pretrained and fine-tuned model highlight the correct area in the map (respectively, alpine regions, urban areas and forest and grassland); in those cases, the fine-tuned models show a stricter and clearer delimitation between land cover and stronger negative values for areas not corresponding to the text input. 
For the maps shown in Figure~\ref{fig:prompts} (f),  the pretrained model map (top) does not react strongly to the prompt (mostly average values, with white colors) and does not recognize the water areas (lakes) from the map, while the task is easily accomplished by the fine-tuned model (bottom map). 

This trend is further confirmed by the examples in Figure~\ref{fig:simi}. We compare the text-image similarity scores for the pretrained and fine-tuned SkyCLIP model on two text-image pairs from the EcoWikiRS dataset and present the top-3 sentences with the highest scores.  Figure~\ref{fig:simi} shows that WINCEL fine-tuning encourages the model to select relevant sentences, with text that is ecologically relevant to the image, such as such as ``sandy", ``urban", ``acidic" and even ``air pollution", possibly due to the proximity to the road. This generalizes to the further examples in the Suppl.  Fig~\ref{fig:simi_suppl}. 
By observing the score values, the fine-tuned model rejects sentences by attributing them a negative score since the text was considered irrelevant. For the pretrained model, the best-ranked sentences are usually related to land cover but not necessarily visually relevant to the associated image (see additional examples in Supp. Section~\ref{blackbird}). This suggests that the original model fails to align the image with the relevant text adequately.

\subsection{Limitations}
While fine-tuning pretrained RS-VLMs on the EcoWikiRS dataset leads to performance gains for EUNIS ecosystem recognition, several limitations arise from merging several resources coming from crowd-sourced platforms. The species observations from GBIF for a given location are incomplete (presence-only surveys), with systematic biases favoring certain taxonomic groups (e.g., butterflies and flowering plants) and introducing seasonal and spatial biases toward some areas (e.g. inhabited areas, parks, etc.)~\cite{daru2018widespread,robertson2014gbif}.  EUNIS ecosystem type maps come with a confidence indicator due to the variety of data sources combined to produce them and might comprehend erroneous labels. Accounting for these biases~\cite{fithian2015bias} could help further improve our approach.
Wikipedia texts are influenced by their contributors and, therefore, the quality and level of detail provided can vary. Using additional text sources could also help improve the learned representation.
The spatial extent of the used imagery is limited to Switzerland. The lack of freely available high-resolution aerial imagery hinders the direct generalization to a larger scale of our proposed approach, as this would require experimenting with other imagery sources with coarser resolution.

%% file: table1.tex
\begin{table}[t]
\centering
\resizebox{0.75\columnwidth}{!}{%
\begin{tabular}{llll}
\toprule
\textbf{Model} & \textbf{Training} & \textbf{OA} & \textbf{F1} \\  \hline
                                     & pretrained & 14.7                           & 12.3                            \\
\textbf{CLIP }                       & infoNCE    & 25.3           $^{\pm0.3} $    & 17.1             $^{\pm0.3}$      \\
                                     & WINCEL     & \textbf{30.9}  $^{\pm0.7} $    & \textbf{20.1}    $^{\pm0.4}$      \\ \hline
                                     & pretrained & 28.5                           & 19.0                            \\
\textbf{GeoRSCLIP}                   & infoNCE    & 27.6           $^{\pm0.2} $    & 19.6             $^{\pm0.3}$         \\
                                     & WINCEL     & \textbf{29.5}  $^{\pm0.2} $    & \textbf{20.4}    $^{\pm0.1}$      \\ \hline
                                     & pretrained & 11.3                           & 7.6                             \\
\textbf{RemoteCLIP}                  & infoNCE    & \textbf{22.2}  $^{\pm0.2} $    & \textbf{13.3}    $^{\pm0.2}$       \\
                                     & WINCEL     & 20.9           $^{\pm0.3} $    & 11.9             $^{\pm0.5}$     \\ \hline
                                     & pretrained & 19.2                           & 13.7                            \\
\textbf{SkyCLIP}                     & infoNCE    & 27.1           $^{\pm0.5} $    & 18.5             $^{\pm0.2}$     \\
                                     & WINCEL     & \textbf{30.1}  $^{\pm0.3} $    & \textbf{20.4}    $^{\pm0.3}$      \\ \hline
\multicolumn{2}{l}{\textbf{Supervised upper-bound}} & 52.7                         & 39.8                            \\ \bottomrule          
\end{tabular}%
}
\caption{Prediction performance on the EUNIS ecosystem labels in zero-shot settings for pretrained RS-VLMs and  models fine-tuned on the EcoWikiRS dataset with ``habitat'' sentences. We compare the InfoNCE loss and the proposed WINCEL. We report mean and standard deviation over 5 runs (except for pretrained model weights). OA = overall accuracy. F1= macro F1-score. 
}\label{tab:table1}
\end{table}

%% file: table2.tex
\begin{table}[t]
\centering
\resizebox{0.6\columnwidth}{!}{
\begin{tabular}{lll}
\toprule
\textbf{Input text}  & \textbf{OA}             & \textbf{F1}  \\ \midrule
habitat  & \textbf{30.1}    $^{\pm0.3 } $        &\textbf{ 20.4}     $^{\pm0.3 }$  \\
keywords & 28.3             $^{\pm0.5 } $        & 19.3              $^{\pm0.4}$  \\
random   & 27.2             $^{\pm0.1 } $        & 18.6              $^{\pm0.5}$  \\
species names  & 19.3       $^{\pm0.2}  $        & 13.8              $^{\pm0.1}$  \\ \bottomrule
\end{tabular}
}
\caption{Ablation study on the type of input sentences extracted from Wikipedia for each species. Results are reported for the SkyCLIP model trained with WINCEL.}\label{tab:Table2}
\end{table}

%% file: table4.tex
\begin{table}[t]
\centering
\resizebox{0.6\columnwidth}{!}{%
\begin{tabular}{lll}
\toprule
 \textbf{Approach} & \textbf{OA }   & \textbf{F1}            \\ \midrule
InfoNCE             & 27.1  $^{\pm0.5 } $          & 18.5           $^{\pm0.2 } $       \\
sampling top-1      & 28.0  $^{\pm0.4 } $          & 19.9           $^{\pm0.5 } $       \\
sampling top-p      & 25.3  $^{\pm0.7 } $          & 16.0           $^{\pm0.7} $          \\
substring augm      & 26.9  $^{\pm 0.3} $          & 17.2           $^{\pm0.4} $          \\ 
bootstrap-hard      & 28.4  $^{\pm0.7 } $          & 19.7           $^{\pm0.7 } $       \\
bootstrap-soft      & 21.4  $^{\pm0.4 } $          & 12.7           $^{\pm0.8 } $       \\  
WINCEL              & \textbf{30.1} $^{\pm0.3 } $  &\textbf{20.4}   $^{\pm0.3 } $       \\
\bottomrule
\end{tabular}%
}
\caption{Comparison of different methods addressing the noisy label issue.  SkyCLIP and ``habitat'' sentences are used in all cases.}\label{tab:table4}
\end{table}

%% file: fig_prompts.tex
\begin{figure*}[ht!]
    \centering
    \subfloat[Annual mean temperature (C\degree ) for Switzerland (1991-2020) \textcopyright MeteoSwiss ]{\includegraphics[width=0.22\textwidth]{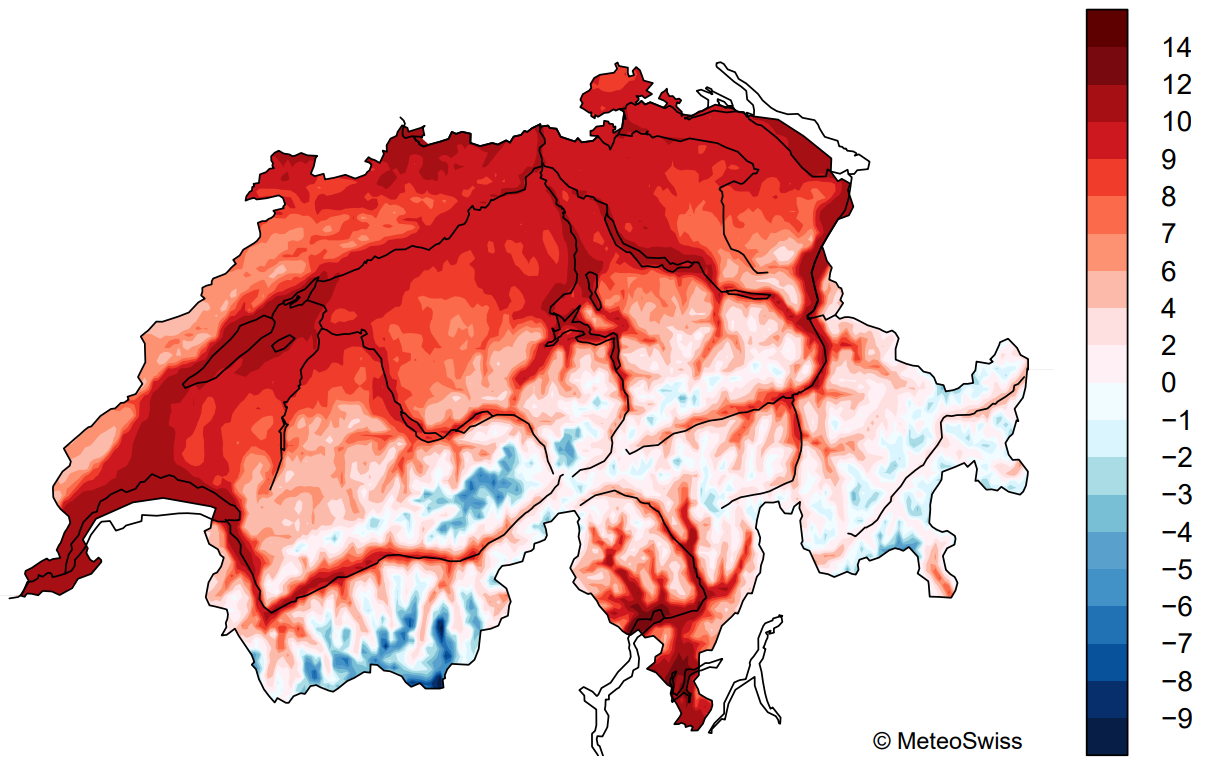}} \quad 
    \quad
    \subfloat[It prefers a warm and dry \\ climate.]{\includegraphics[width=0.22\textwidth]{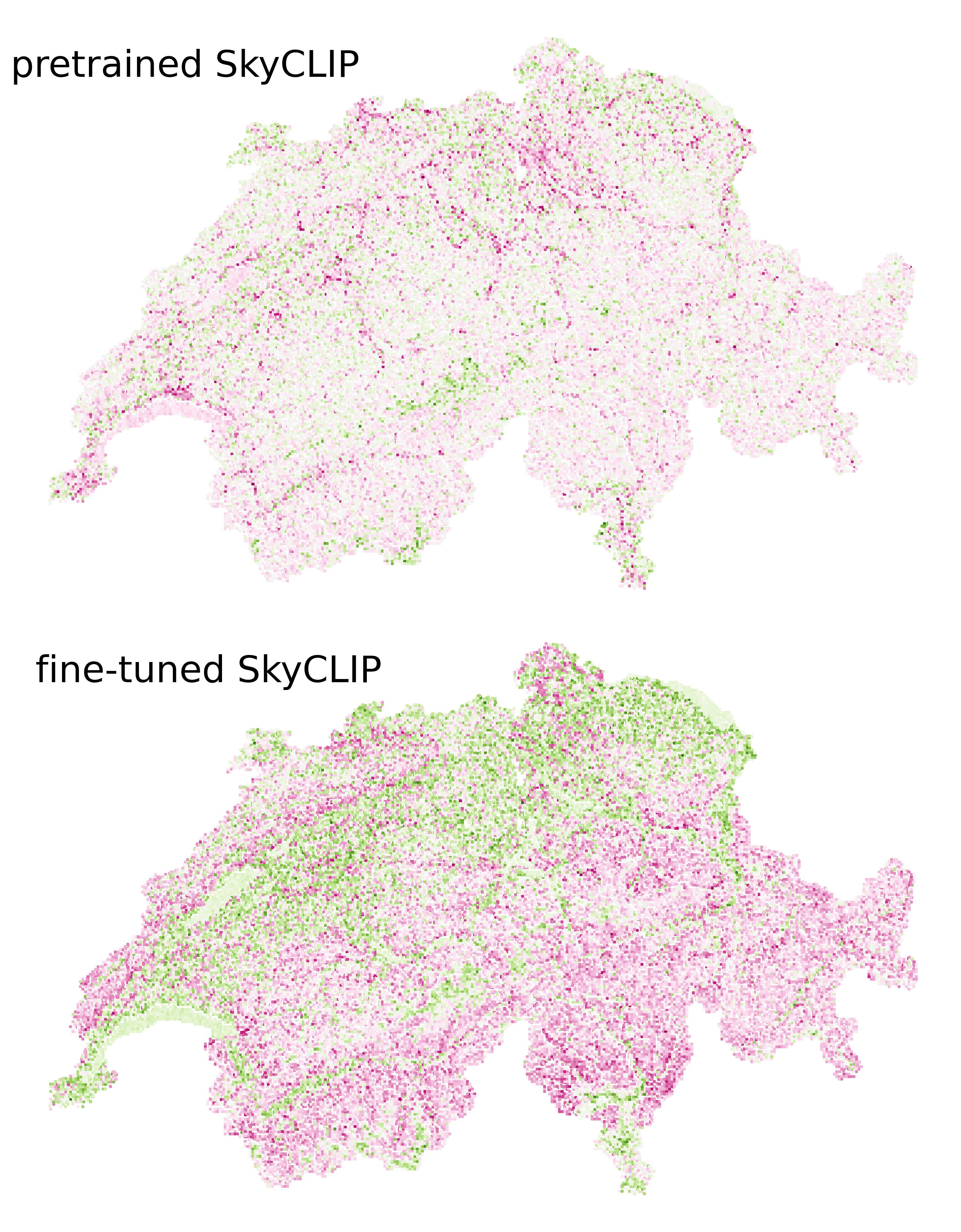}} 
    \quad
    \subfloat[ It occurs on poorly drained neutral and acidic soils of the lowlands and upland fringe.]{\includegraphics[width=0.22\textwidth]{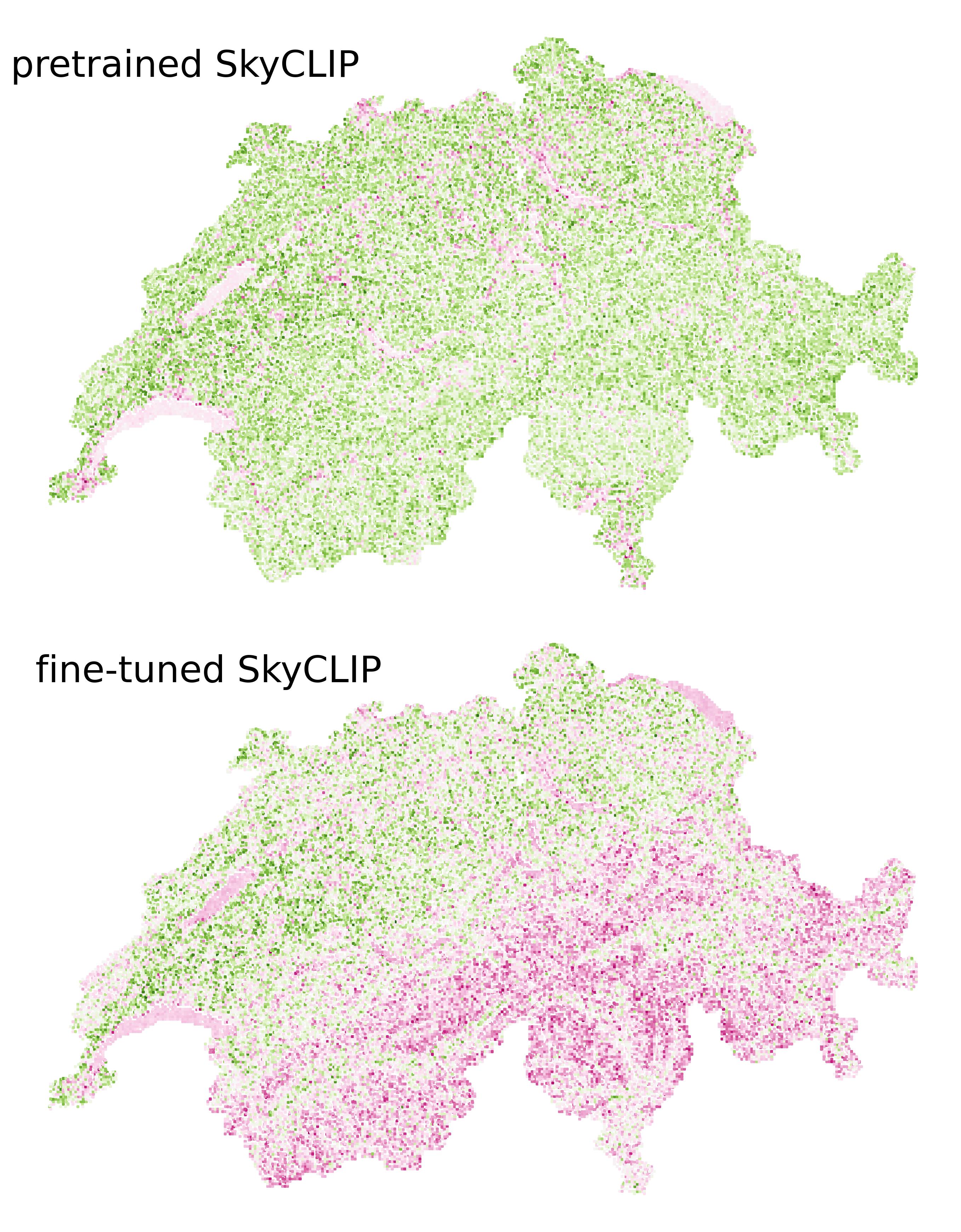}}
    \quad
    \subfloat[It prefers the alpine climatic zone.]{\includegraphics[width=0.22\textwidth]{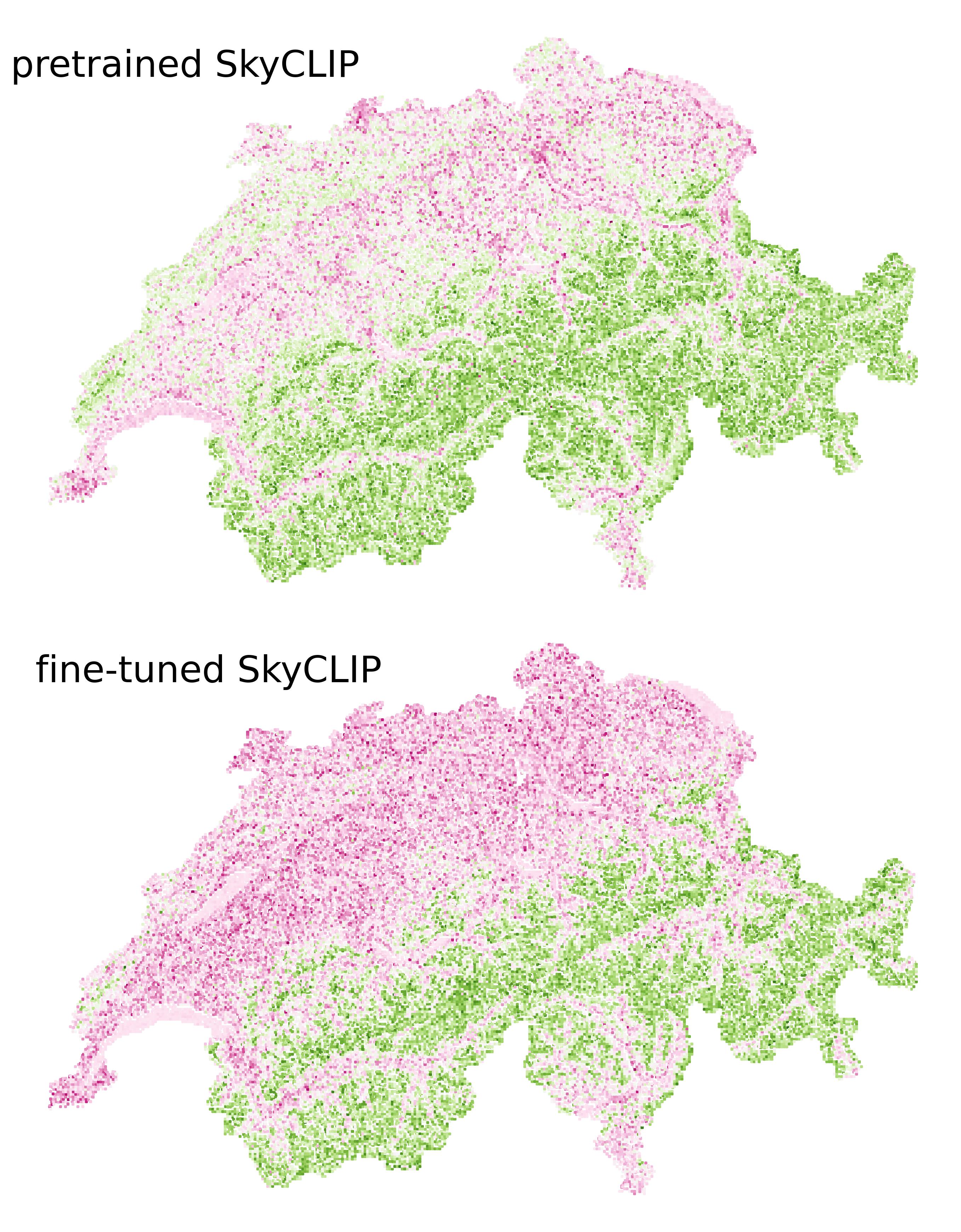}} 
    \\
    \subfloat[Simplified land cover map of Switzerland with EUNIS labels at level L1]{\includegraphics[width=0.23\textwidth]{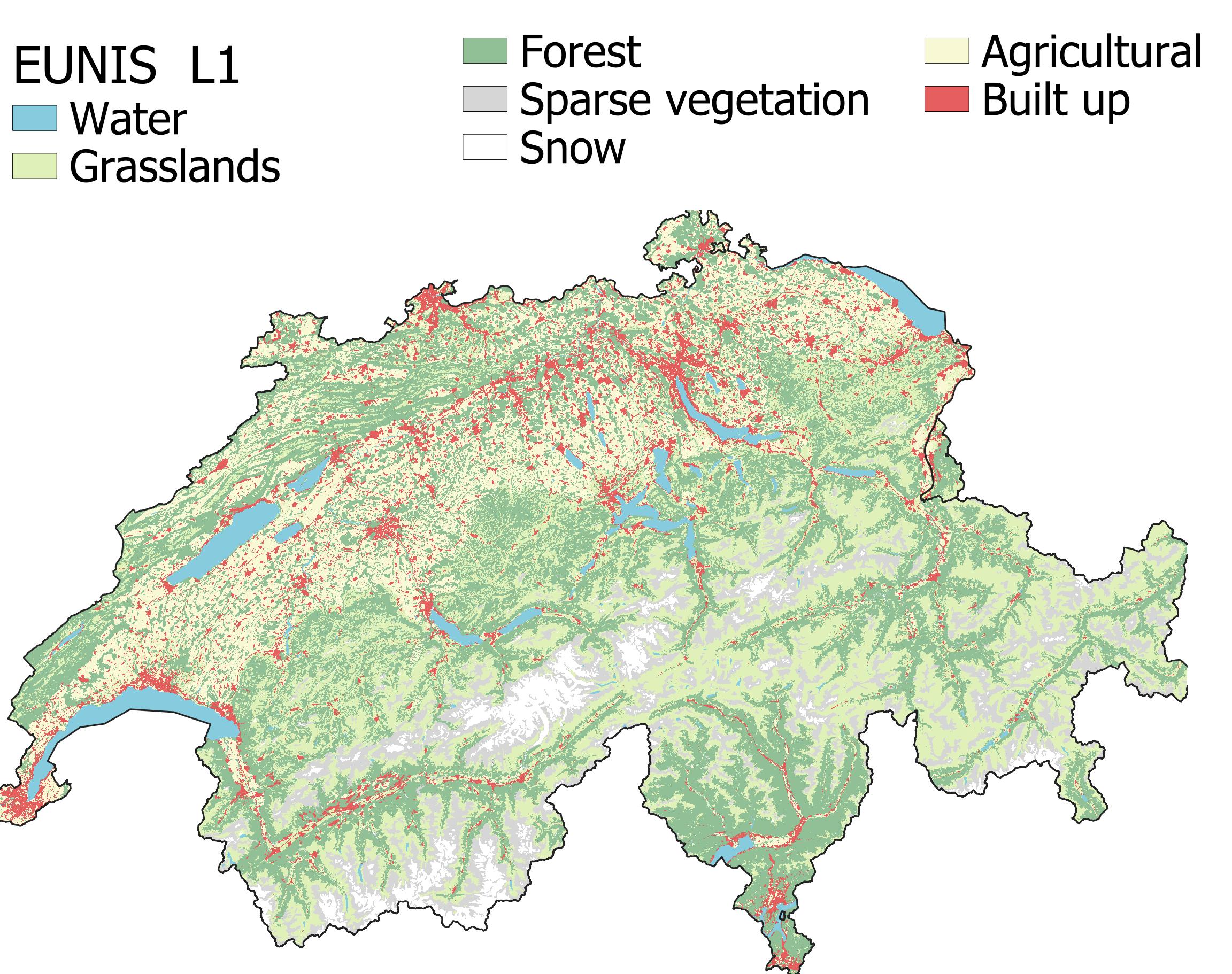}} 
    \quad    
    \subfloat[It is found in both fresh- and salt-water wetlands, including parks, small ponds, rivers, lakes and estuaries.]{\includegraphics[width=0.22\textwidth]{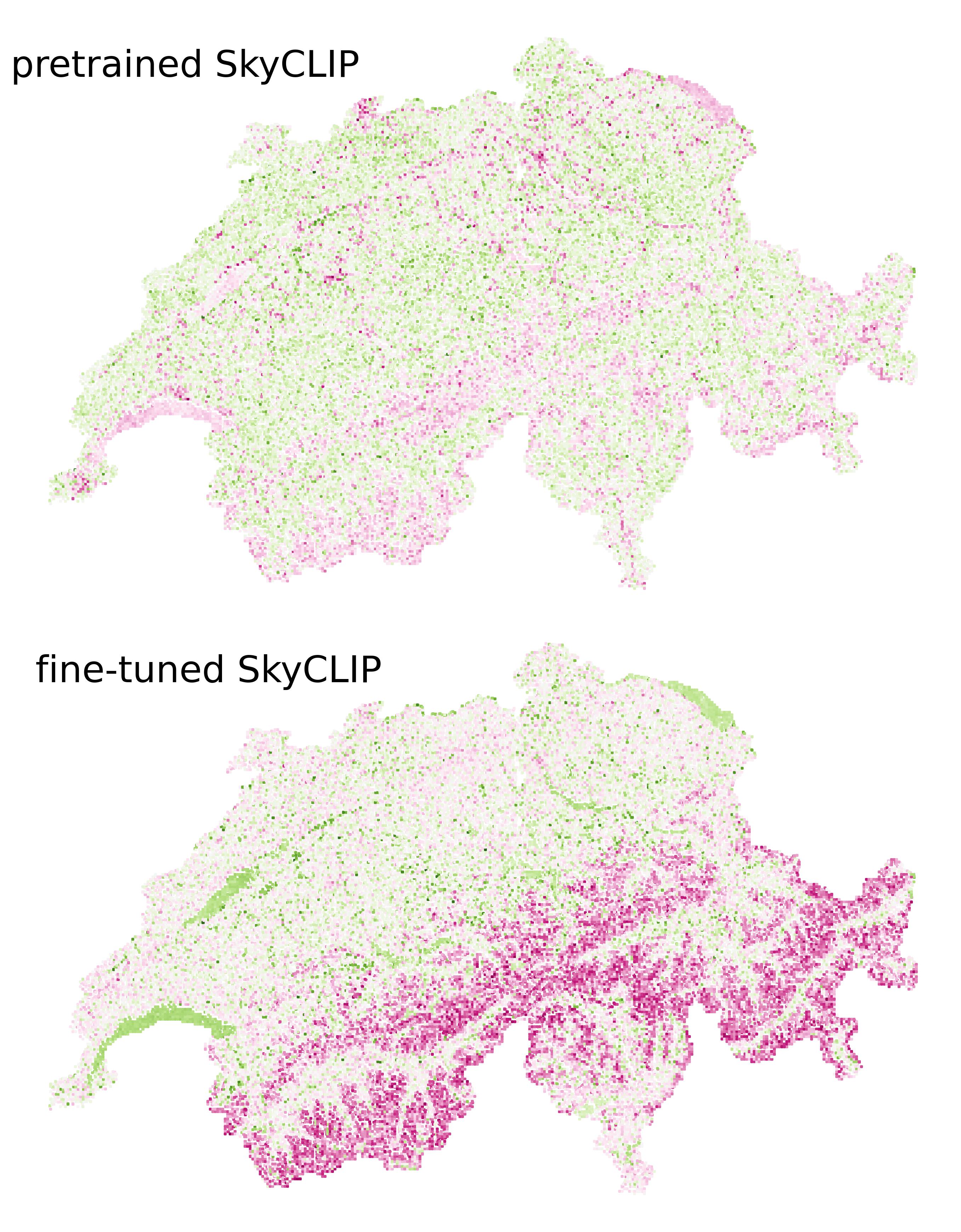}}
    \quad
    \subfloat[While urban red foxes will scavenge successfully in the city some urban residents will deliberately leave food out for the animals, finding them endearing.]{\includegraphics[width=0.22\textwidth]{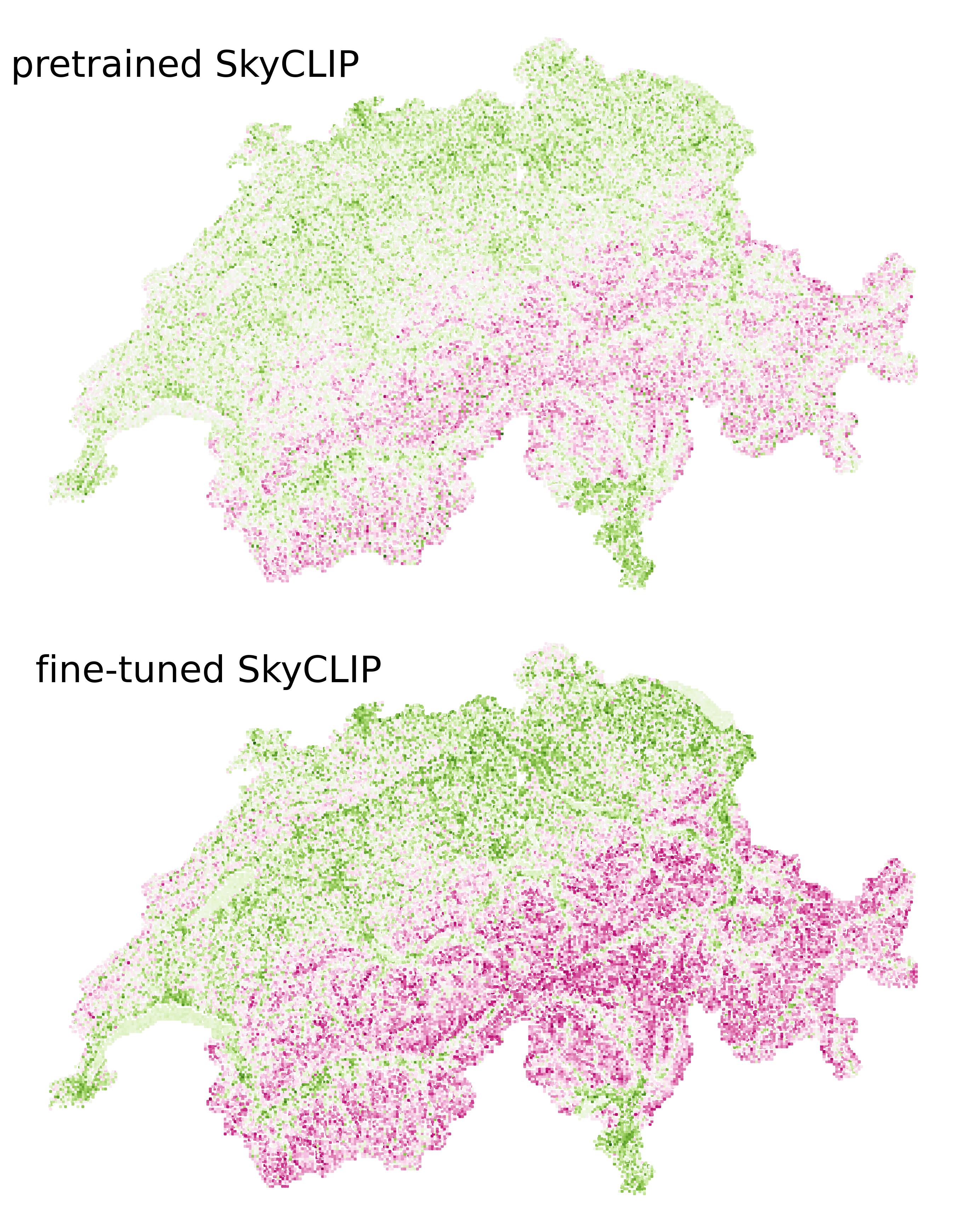}}
    \quad
    \subfloat[It is a long-lived tree of high-canopy woodland, coppice and wood pasture, and it is commonly planted in hedges.]{\includegraphics[width=0.22\textwidth]{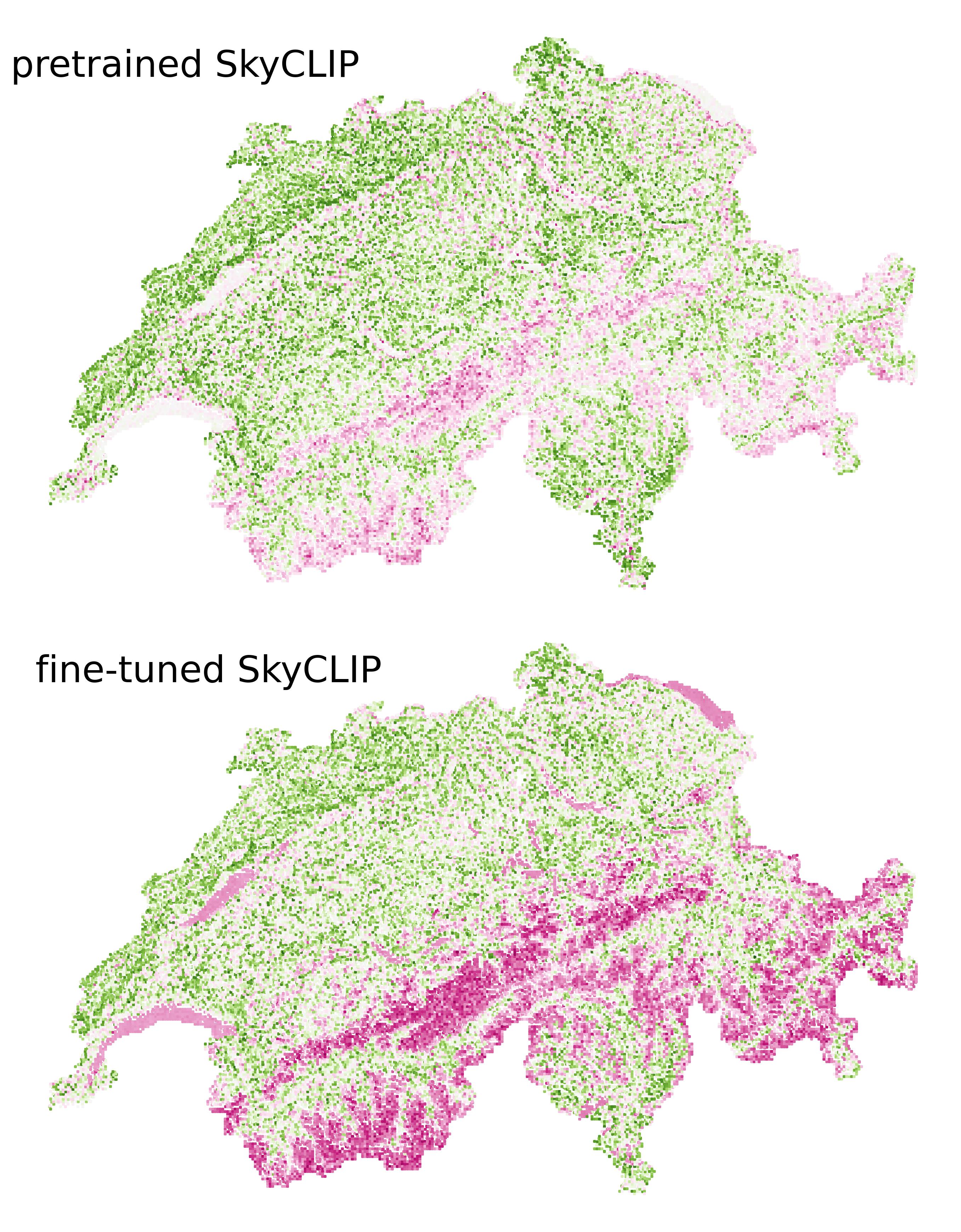}}
    \caption{Visualization of text-image similarity values over the surface of Switzerland with different text prompts as inputs. The  maps uses pretrained SkyCLIP at the top and  fine-tuned  SkyCLIP at the bottom. The simplified land cover map and the temperature maps are for reference. \textcolor{OliveGreen}{High similarity values} are shown in green, while magenta depicts \textcolor{magenta}{lower values} with min-max scaling. Best viewed in colors. }    
    \label{fig:prompts}
\end{figure*}

%% file: table_simi.tex
\begin{figure*}[t]
    \centering
    \begin{minipage}{0.15\textwidth}  
        \centering
        \includegraphics[width=\linewidth]{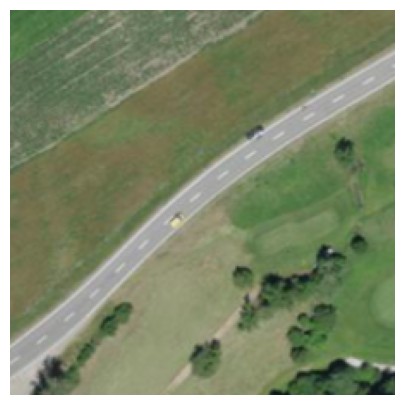} 
    \end{minipage}
    \begin{minipage}{0.84\textwidth} 
        \footnotesize
        \begin{tabular}{rl} 
            \textbf{Score} & \textbf{Pretrained SkyCLIP model } \\ 
            0.210 &S. aucuparia appears north of the boreal forest at the arctic tree line, in Norway, it is found up to the 71st parallel north.\\
            0.193  &In Central Europe it often grows in association with red elderberry, goat willow, Eurasian aspen, and silver birch. \\
            0.191  &The species was introduced as an ornamental species in North America.\vspace{0.1cm}  \\
            \textbf{Score} & \textbf{SkyCLIP model fine-tuned with WINCEL}  \\ 
            0.129  &It mostly grows on soil that is moderately dry to moderately damp, acidic, low on nutrients, sandy, and loose.\\
            0.121  &The plant is also resistant to air pollution, wind, and snow pressure.\\
            -0.039  &The species was introduced as an ornamental species in North America.\\
        \end{tabular}
    \end{minipage}
    \begin{minipage}{0.15\textwidth}  
        \centering
        \includegraphics[width=\linewidth]{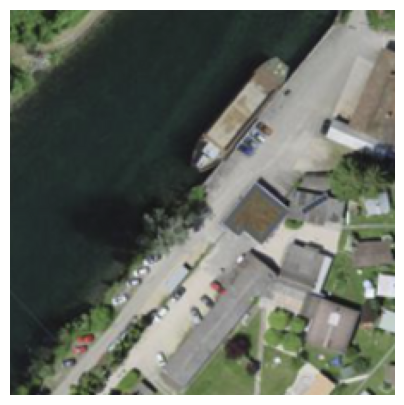} 
        \noindent
    \end{minipage} 
    \begin{minipage}{0.84\textwidth} 
        \footnotesize
            \vspace{0.2 cm}
        \begin{tabular}{rl}            
           \textbf{Score} & \textbf{Pretrained SkyCLIP model } \\ 
            0.197 & The common blackbird occurs at elevations up to 2000m. in Europe, in North Africa and in peninsular India.\\ 
            0.193& The common blackbird also lives in parks, gardens and hedgerows.\\ 
            0.189& Urban males are more likely to overwinter in cooler climes than rural males, an adaptation made   	feasible [...]  \vspace{0.1cm}\\ 
             
            \textbf{Score} & \textbf{SkyCLIP model fine-tuned with WINCEL}  \\ 
            0.192 &  In urban and suburban environments with high levels of anthropogenic noise, such as near airports, blackbirds [...]\\ 
            -0.046 & The common blackbird also lives in parks, gardens and hedgerows.\\ 
            -0.050 & Pairs stay in their territory throughout the year where the climate is sufficiently temperate.\\             
        \end{tabular}
    \end{minipage}
    
    \caption{Comparison of top-3 sentences scores given by the pretrained and fine-tuned SkyCLIP model on  samples of the WikiRS dataset. Scores are cosine similarity values between the text and images features. Scores are ranked by decreasing order of magnitude.   More examples are available in the Suppl. Figure~\ref{fig:simi_suppl}.}
    \label{fig:simi}
\end{figure*}

%% file: 6_conclusion.tex
\section{Conclusion}
\label{sec:conclusion}
In this paper, we propose to learn an ecologically rich embedding space between remote sensing images and texts that describe the local environmental conditions. To do so, we learn text embeddings from Wikipedia articles describing species' habitats. 
For this task, we introduce the EcoWikiRS dataset, connecting high-resolution aerial images with species observations and the textual description of their habitats in Wikipedia. 
To learn from such a dataset, we designed the WINCEL loss to better learn from noisy and weak supervision. 
We evaluate our model on the task of EUNIS ecosystem zero-shot mapping. 
Our approach outperforms pre-trained models and standard methods designed to handle noisy datasets. 
We demonstrated that our framework can also effectively generate high-quality zero-shot maps from fine-grained text prompts to highlight environmental properties at the country scale, and that it allows to select sentences that are relevant to a RS image in a more convincing manner than a pretrained RS-VLM.  We hope that this study paves the way for learning better representations of RS images integrating ecological concepts.

%% file: 7_suppl.tex
\maketitlesupplementary 
\input{table10_black_bird}
\input{table6}
\input{table_text_stats}
\input{table5}

\section{Study of fine-tuning strategies.}
Table~\ref{tab:table6} compares the performance of SkyCLIP under various fine-tuning configurations with WINCEL.  The text encoder remains frozen, while different layers of the visual encoder undergo fine-tuning: the final transformer block, the last projection layer with or without additional projection layers, and fine-tuning with  Exponential Moving Average (EMA). By fine-tuning the last transformer block,  the performance significantly deteriorates compared to the other strategies. This may result from the loss of original pretrained representations, as the fine-tuning dataset is comparatively small. Both EMA fine-tuning and updating the last projection layer obtain good performances, while being inferior to training the positional encoding and the last projection layer. The latter outperforms all other fine-tuning strategies, both in terms of OA and F1 score. Therefore, we adopted this strategy in this work.  
\section{Additional visual results} \label{blackbird}
As discussed in Section~\ref{sec:method}, generalist species, such as the common blackbird, can live in several habitats. Thus, sentences describing their habitat are likely to be uninformative or irrelevant when paired with a given image. 
To demonstrate that our model accurately identifies sentences relevant to a given image, we study the cross-modal similarity scores between sentences describing the common blackbird habitat, a generalist species, with aerial images depicting different land covers.  We computed the text and visual representations for $4$ aerial images and $6$ sentences from the Wikipedia article of the \textit{Turdus merula} (common blackbird) with  SkyCLIP model fine-tuned with the WINCEL loss and present the cross-modal similarity scores in Figure~\ref{fig:black_bird}. When the image (ex. image (a))  or the text (sentences 5 and 6) is irrelevant to the other modality, the similarity scores are close to zero or even negative. Conversely, pertinent image-text pairs obtain high similarity values. 
\section{Prompting for zero-shot classification}
Due to the presence of spurious concept biases, several works highlighted the importance of prompt engineering~\cite{radford2021learning,allingham2023simple} for attaining high zero-shot classification performances. To select satisfying prompts for the EUNIS ecosystem classification task, we compared manually designed prompt templates.  The results are shown in Table~\ref{tab:Table5}. Overall, the prompt with EUNIS ecosystem names obtains the highest performance for all backbones, except for RemoteCLIP, whose performances significantly increase by using a prompt template such as "a remote sensing image of \{\}".
Figure~\ref{fig:simi_suppl} presents additional illustrations of cross-modal similarity values, similar to those presented in Figure~\ref{fig:simi},  with the top-5 most similar sentences for the pretrained and the fine-tuned SkyCLIP model for each dataset sample.

\section{Dataset construction and statistics}

\subsection{Wikipedia articles parsing}

\textbf{Extraction of Wikipedia articles.} 
We downloaded a dump of all of Wikipedia through Wikimedia at \url{dumps.wikimedia.org}. We processed the dump with the \textit{BeautifulSoup}~\cite{richardson2007beautiful} and \textit{mwparserfromhell} python packages. We extracted articles matching the ``Speciesbox" template. We created the species binomial name using the genus and the species string from the  Speciesbox. We matched the species article to GBIF species and saved the article content. We cut the articles into sections and extracted sentences from the habitat section, containing keywords from the list mentioned below and also saved all sentences. Statistics on the total number of sentences and unique sentences are shown in Table~\ref{tab:sentences_count}.
\\

\noindent\textbf{Extraction of keywords sentences.} \label{sec:keywords}
We selected sentences containing at least one of the following strings:
\footnotesize
urban, city, town, road, railway, rail, highway, port, airport, mineral, dump, construction, green, sport, arable, farmland, irrigated, fruit, berry, plant, tree, olive, crop, pastures, vineyards, cultivation, agriculture, vegetation, forest, forestry, grassland, heathland, moors, woodland, shrub, beach, dunes, sand, rock, bareland, vegetated, inland, marshes, burnt, water, coast, coastal, lagoons, sea, ocean, saline, peatbogs, estuaries, surface, grass, grassland, dry, mesic, littoral, seasonal, wet, alpine, subapline, arctic, scrub, temperate, temperature, mediterranean-montane, plantation, coniferous, deciduous, anthropogenic, coppice, screes, inland, cliffs, outcrops, snow, ice, ice-dominated, garden, park, village, building, transport, hard-surfaced, constructed, runway, airport, road, vehicle, bridge, shrubwood, weed, bareland, fanshaped, ravine, gravel, rectangular, high, low, coastline, cemetery, greenbelt, circular, cloud, dam, terrace, weed, viaduct, wetland, wood, habitat, ecosystem, landcover, eco, supralittoral, zone, area, saline, density, arborescent, hot, cold, thermo, warm, xerophytic, calcareous, broadleaved, leave, mires, pavements, shores, salt, montane, polygon, evergreen, waste, sparse, dense, coastal, atlantic, anthropogenic, reed, shingle, mediterranean, artificial, park, flower, prairie.

\subsection{GBIF filtering}\label{app:gbif_filtering}
The GBIF download~\cite{gbif_download} included the following criteria: 
\begin{itemize}
    \item \textit{BasisOfRecord }is one of (Observation, Machine Observation, Human Observation, Living Specimen, Occurrence evidence)
    \item \textit{Country} is Switzerland
    \item \textit{GadmGid} is CHE
    \item \textit{TaxonKey} is one of (Animalia, Plantae)
    \item \textit{Year} 1950-2024
\end{itemize}
The observations matching one of the following filtering criteria  were removed in Python :
\begin{itemize}
    \item \textit{coordinateUncertaintyInMeters} is larger than $ 100 m$
    \item \textit{coordinateUncertaintyInMeters} is None
    \item \textit{species} is None
    \item \textit{issue flags} is COORDINATE ROUNDED
    \item The species does not have an article in Wikipedia with a habitat section, based on the species binomial name.
\end{itemize} 
We additionally removed duplicates (multiple records of a species from a particular location).

\begin{figure*}[ht]
    \centering
    \includegraphics[width=\linewidth]{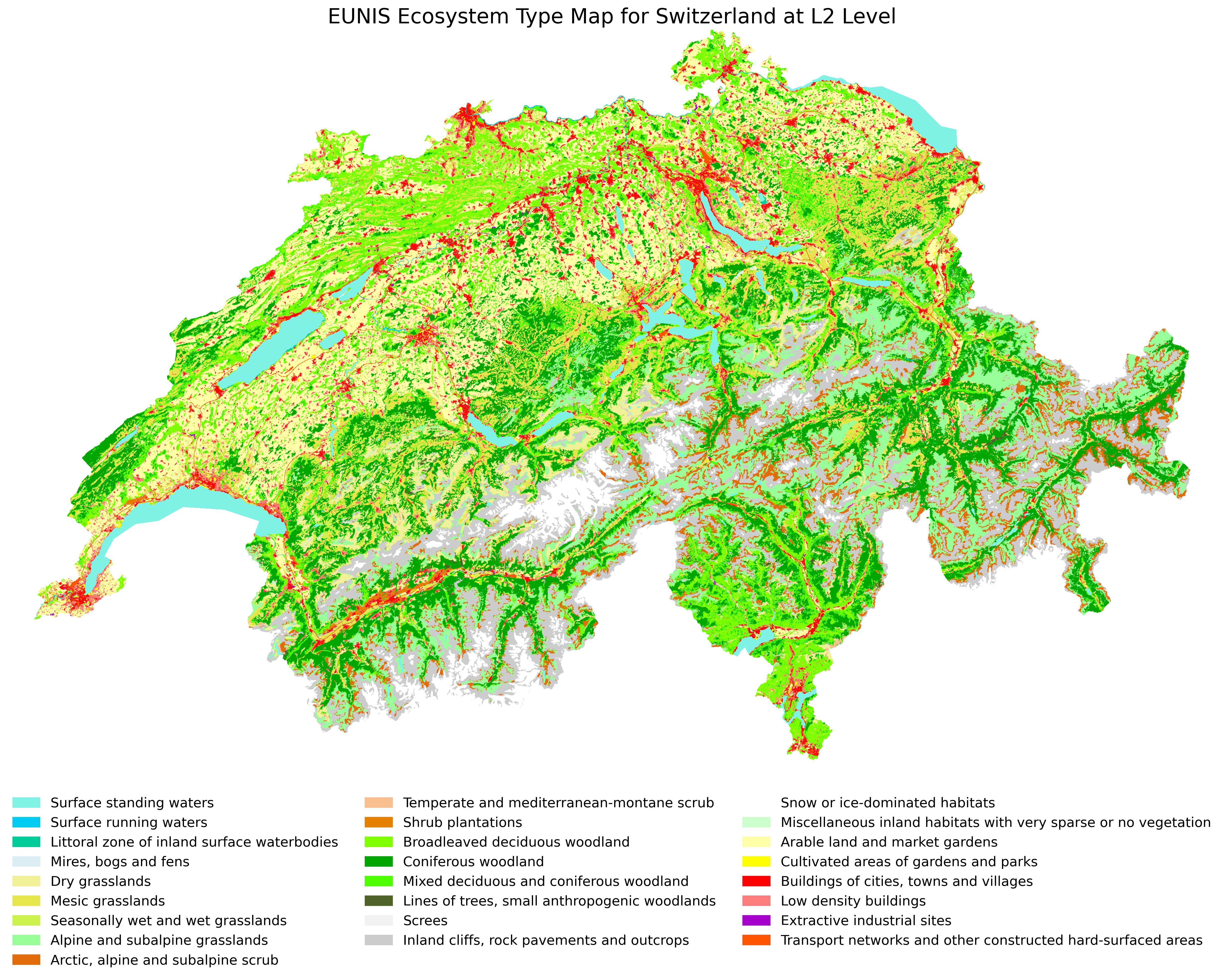}
    \caption{The EUNIS ecosystem type map in Switzerland at level L2.}
    \label{fig:eunis_habitatl}
\end{figure*} 
\begin{figure}[ht]
\centering
    \centering
    \includegraphics[width=0.9\linewidth]{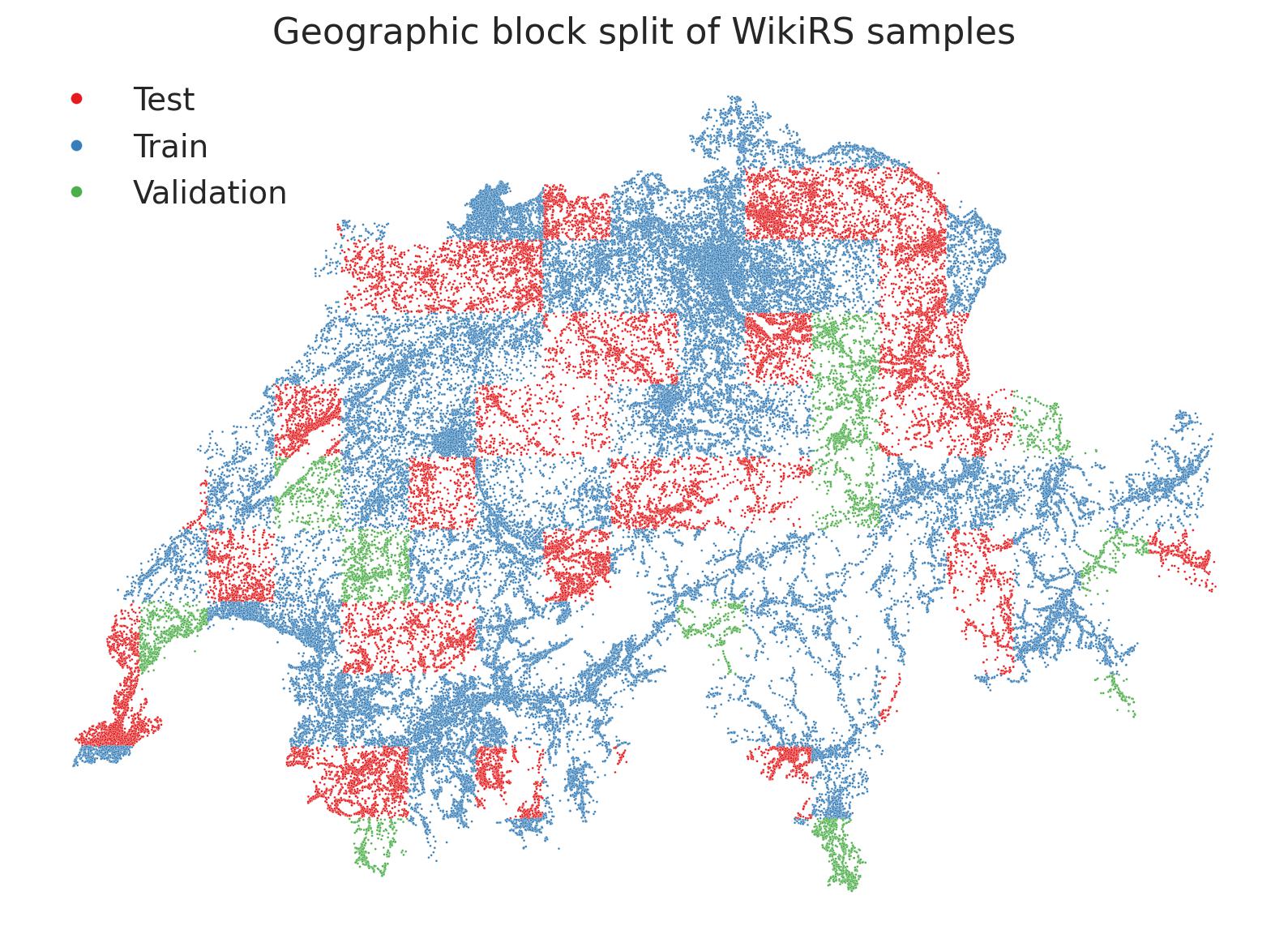}
    \caption{Distribution of our training samples across Switzerland between training, testing and validation sets following a block split approach (with a size of $20$ km).}
    \label{fig:block_split}
\end{figure}
\begin{figure}[ht]
\centering
    \centering
    \includegraphics[width=\linewidth]{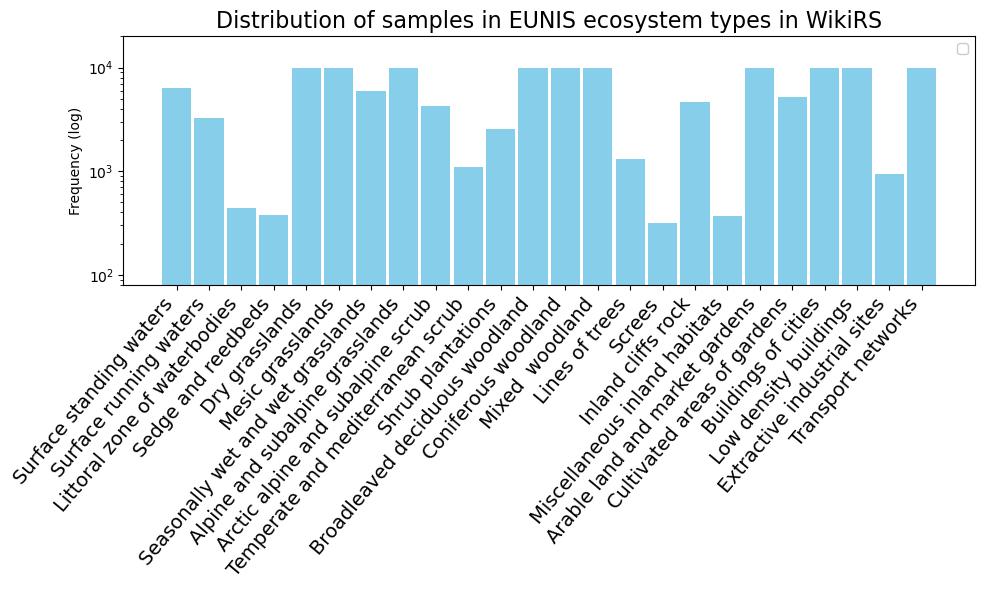}
    \caption{Distribution into EUNIS ecosystem types of samples from the WikiRS dataset on a log scale}
    \label{fig:barplot_distr_eunis}
\end{figure} 
\input{table_simi_suppl}

\input{table9_dataset_samples}

%% file: table10_black_bird.tex
\begin{figure*}[ht]
\begin{minipage}{0.25\textwidth}
\centering
\includegraphics[width=\linewidth]{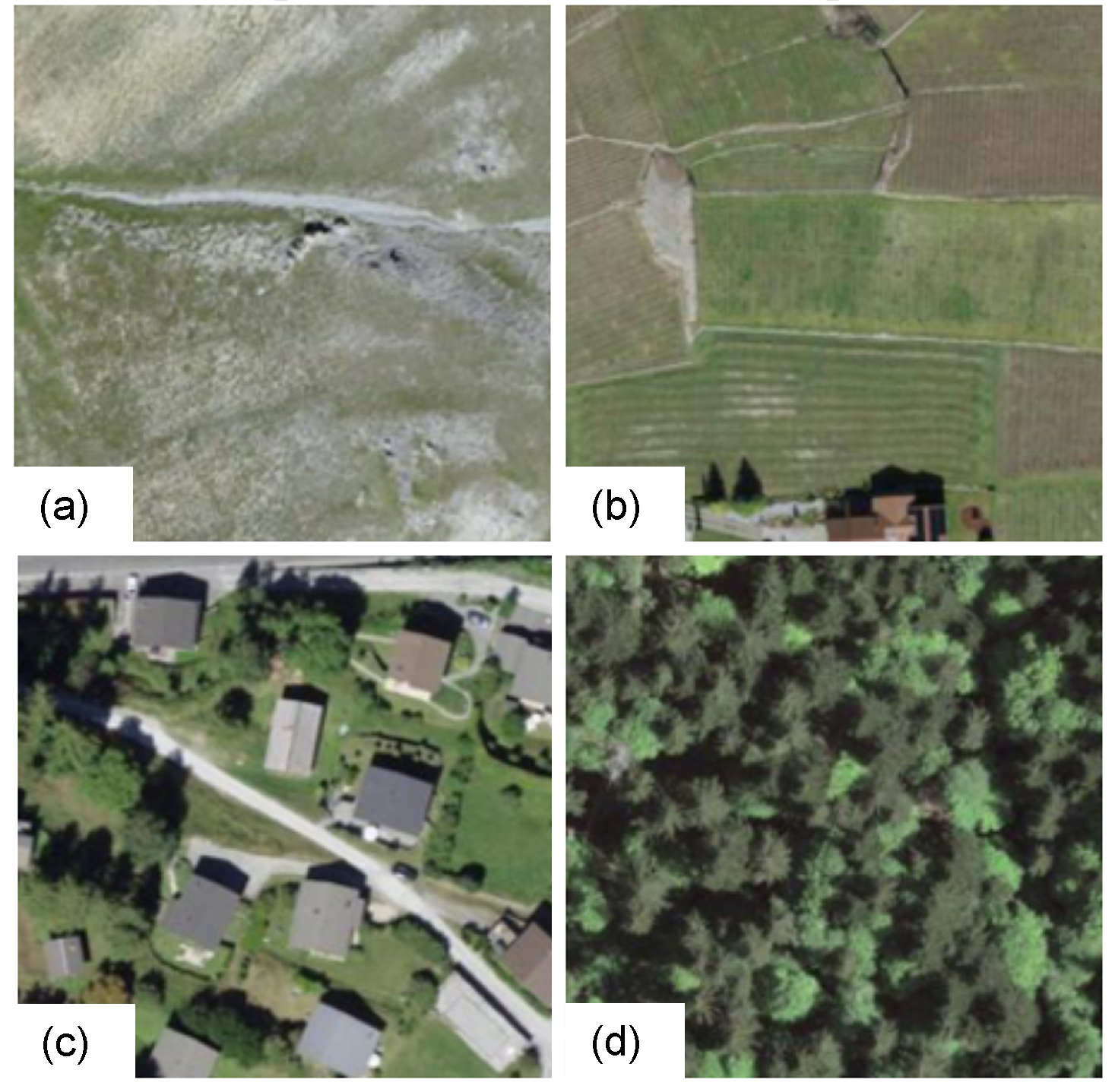}
\end{minipage}
\begin{minipage}{0.75\textwidth}
\centering
\resizebox{\textwidth}{!}{%
%
\begin{tabular}{m{0.8cm} m{0.8cm} m{0.8cm} m{0.8cm} m{11cm}}
\textbf{(a)} &\textbf{(b)} & \textbf{(c)} & \textbf{(d)} &
\textbf{Sentences from the Wikipedia article of \textit{Turdus merula} } \\ \midrule
%
%
%

%
-0.66 &
0.19 &
0.40 &
0.05 &
The common blackbird also lives in parks, gardens and hedgerows.    \vspace{0.4cm}\\

-0.64 &
0.06 &
0.42 &
0.73 &
Common over most of its range in woodland, the common blackbird has a preference for deciduous trees with dense undergrowth. \\

-0.49 &
0.54 &
0.21 &
0.16 &
It eats a wide range of native and exotic fruit, and makes a major contribution to the development of communities of naturalised woody weeds. \\

-0.47 &
-0.06 &
0.24 &
-0.14 &
Near human habitation, the main predator of the common blackbird is the domestic cat, with newly fledged young especially vulnerable. \\
-0.37 &
-0.01 &
-0.03 &
-0.19 &
The common blackbird breeds in temperate Eurasia, North Africa, the Canary Islands, and South Asia. \vspace{0.01cm}\\
-0.25 &
-0.26 &
-0.23 &
0.00 &
Pairs stay in their territory throughout the year where the climate is sufficiently temperate.  \\ 
\end{tabular}%
}
\label{tab:turdula_merlu}
\end{minipage}
\caption{Cross-modal similarities values between sentences from the Wikipedia article of   \textit{Turdus merula} (common black bird) and various images from our dataset representing different land cover. \textcolor{OliveGreen}{High similarity values} are shown in green, and \textcolor{red}{low similarity values} are depicted in red. }
\label{fig:black_bird}
\end{figure*}

%% file: table6.tex
\begin{table}[ht]
\begin{tabular}{p{4cm}p{1cm}p{1cm}}
\toprule
Fine-tuning strategy    & OA    & F1    \\ \midrule
Last transformer block           & 21.6            & 10.8         \\
Last projection layer            & 28.5            & 18.8          \\
EMA                              & 28.9            & 20.1          \\
\multirow{2}{4cm}{Positional encoding \\ + last projection layer (our)} & \multirow{2}{1cm}{\textbf{30.1}} &  \multirow{2}{1cm}{\textbf{20.4}} \\ 
&& \\ \bottomrule
\end{tabular}
\caption{Comparison of fine-tuning strategies for  the SkyCLIP model trained with WINCEL on the EcoWikiRs dataset}
\label{tab:table6}
\end{table}

%% file: table_text_stats.tex
\begin{table}[t]
\centering
\begin{tabular}{p{1.4cm}p{1.7cm}p{1.6cm}p{1.5cm}}
\toprule
 Text type &  \parbox[t]{1.5cm}{  Number of \\ sentences} &   Number of unique sentences & \footnotesize Average number of sentences per location \\
\midrule
habitat  & 2'998'305  & 18'693  & 10.9 \\
keywords & 3'728'644  & 21'832  & 13.6 \\
random   & 19'642'735 & 103'065 & 71.6 \\ 
\bottomrule
\end{tabular}%
\caption{Number of sentences and unique sentences in the different text types of WikiRS dataset}
\label{tab:sentences_count}
\end{table}

%% file: table5.tex
\begin{table*}[th]
\centering
    \begin{tabular}{lllllllll|ll}
    \toprule
     \textbf{Model} &
      \multicolumn{2}{c}{ \textbf{SkyCLIP}} &
      \multicolumn{2}{c}{ \textbf{CLIP}} &
      \multicolumn{2}{c}{ \textbf{GeoRSCLIP}} &
      \multicolumn{2}{c}{ \textbf{RemoteCLIP}} &
      \multicolumn{2}{c}{ \textbf{Average}} \\ \midrule
    \textbf{Prompting approach} &
      \textbf{OA} &
      \textbf{F1} &
      \textbf{OA} &
      \textbf{F1} &
      \textbf{OA} &
      \textbf{F1} &
      \textbf{OA} &
      \textbf{F1} &
      \textbf{OA} &
      \textbf{F1} \\ \midrule
    \textbf{EUNIS class name}                     & \textbf{30.1 }& \textbf{20.4} & \textbf{30.9} & 20.1 & \textbf{29.5} & \textbf{20.4} & 20.9 & 11.9 & \textbf{30.0} & \textbf{20.3 }\\
    \textbf{EUNIS class description}              & 21.8 & 15.4 & 24.7 & 16.5 & 21.8 & 15.4 & 25.0 & 16.0 & 23.3 & 15.8 \\
    \textbf{``an aerial image of \{\}''}        & 28.3 & 19.3 & 29.5 & 20.5 & 28.3 & 19.3 & 24.5 & 17.5 & 27.6 & 19.1 \\
    \textbf{``a remote sensing image of \{\}''} & 29.0 & 18.7 & 30.5 & \textbf{20.9} & 29.0 & 18.7 & \textbf{25.8} & \textbf{18.5} & 28.6 & 19.2 \\
    \textbf{``a satellite image of \{\}''}      & 28.4 & 18.4 & 29.0 & 20.5 & 28.4 & 18.4 & 25.1 & 17.2 & 27.7 & 18.6 \\ \midrule
    \end{tabular}
    \caption{Comparison of different prompts for the pretrained models and those trained with the WINCEL loss on the EcoWikiRS dataset.  OA and F1 values are average metrics values over 5 models, similarly to results from Table~\ref{tab:table1}. }
    \label{tab:Table5}
\end{table*}

%% file: table_simi_suppl.tex
\begin{figure*}[t]
    \begin{minipage}{0.15\textwidth}  
        \centering
        \includegraphics[width=\linewidth]{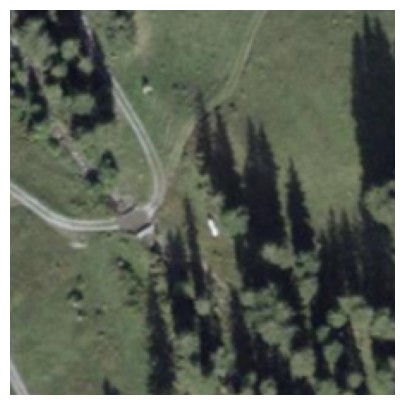} 
    \end{minipage}
    \begin{minipage}{0.84\textwidth} 
        \footnotesize
        \begin{tabular}{ll}
           \textbf{Score} & \textbf{Pretrained SkyCLIP model}  \\ 
            0.233 & It is becoming rarer, particularly in the north of its distribution, largely due to increasingly  intensive agriculture and\\
            & commercial wildcrafting  \\
            0.220 & It is rare overall, but may be locally abundant. \\
            0.218 & Nevertheless, it is cultivated on a large scale in Estonia.\\
            0.213 & Arnica montana grows in nutrient-poor siliceous meadows or clay soils.\\
            0.211 & Arnica montana is widespread across most of Europe.  \\ 
            \\
            \textbf{Score} & \textbf{Fine-tuned SkyCLIP model}  \\ 
            0.657 &Arnica montana grows in nutrient-poor siliceous meadows or clay soils.\\
            0.527 &Arnica montana is widespread across most of Europe.\\
            0.033 &However Arnica does not grow on lime soil, thus it is an extremely reliable bioindicator for nutrient poor and acidic soils.\\
            -0.105 & It is becoming rarer, particularly in the north of its distribution, largely due to increasingly intensive agriculture and \\ & commercial wildcrafting .\\
            -0.130 & It is rare overall, but may be locally abundant.\\
            \\
           \bottomrule
            \\ 
        \end{tabular}
    \end{minipage}
    \begin{minipage}{0.15\textwidth}  
        \centering
        \includegraphics[width=\linewidth]{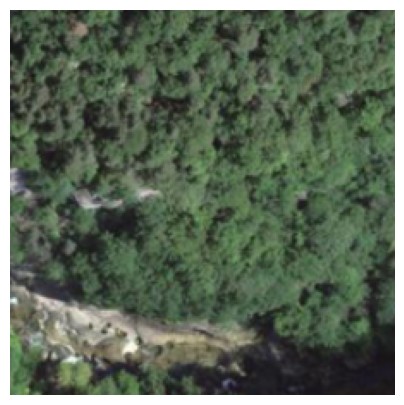} 
    \end{minipage}
    \begin{minipage}{0.84\textwidth} 
        \footnotesize
        \begin{tabular}{ll}   
            \textbf{Score}& \textbf{Pretrained SkyCLIP model }\\ 
            0.265 &This species was once abundant, when forests were used for grazing livestock and trees were coppiced, but is now\\ 
            & threatened by overgrowth of larger plants.\\
            0.25& Bjerkandera adusta, Phlebia acerina, Sebacinaceae, Tetracladium sp., and Tomentella sp.  Cephelanthera longifolia \\ 
            &is vulnerable to grazing by deer.\\
            0.248& Cephalanthera longifolia is common in some parts of its European range, such as southern France and Spain, but\\ 
            & endangered particularly in northern areas such as Belgium.\\
            0.237& In 2007 it was listed as a priority species under the UK Biodiversity Action Plan.\\
            0.231 &Another investigation indicated 9 mycorrhizal partners .\\
            \\
            \textbf{Score}& \textbf{Fine-tuned SkyCLIP model}  \\ 
            0.803 &Sword-leaved helleborine usually grows in damp woodland places , forest edges and rocky slopes.\\
            0.639 &Cephalanthera longifolia is common in some parts of its European range, such as southern France and Spain, but\\ 
            & endangered particularly in northern areas such as Belgium.\\
            0.549& Bjerkandera adusta, Phlebia acerina, Sebacinaceae, Tetracladium sp., and Tomentella sp.  Cephelanthera longifolia\\ 
            & is vulnerable to grazing by deer.\\
            0.309& This species was once abundant, when forests were used for grazing livestock and trees were coppiced, but is now \\ 
            &threatened by overgrowth of larger plants.\\
            0.051& In 2007 it was listed as a priority species under the UK Biodiversity Action Plan.\\
            \\
           \bottomrule
            \\ 
        \end{tabular}
    \end{minipage}    
\begin{minipage}{0.15\textwidth}  
    \centering
    \includegraphics[width=\linewidth]{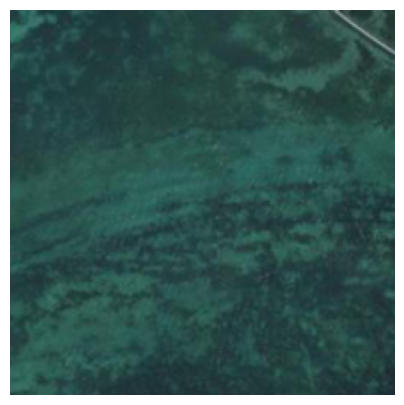} 
\end{minipage}
\begin{minipage}{0.84\textwidth} 
    \footnotesize
    \begin{tabular}{ll}      
        \textbf{Score}& \textbf{Pretrained SkyCLIP model }\\ 
        0.250 &  These dense concentrations of birds are thought to be a defense against attacks by birds of prey such as \\ 
            & peregrine falcons or Eurasian sparrowhawks.\\
        0.240 &  Many birds remain in the same area all year round, but others migrate to spend the winter in mild areas of \\ 
            & western Europe or head south as far as Senegal, Gambia and the Red Sea.\\
        0.237 & Common starlings in the south and west of Europe and south of  latitude 40N are mainly resident.\\ 
            & winter in the southwest of the US.\\ 
        0.236 &  The Eurasian coot is much less secretive than most of the rail family, and can be seen swimming on open water\\ 
            &  or walking across waterside grasslands.\\
        0.235 &  It bobs its head as it swims, and makes short dives from a little jump.\\
        \\         
        \textbf{Score}& \textbf{Fine-tuned SkyCLIP model}  \\ 
       0.896 &  It bobs its head as it swims, and makes short dives from a little jump.\\
        0.682 &  The Eurasian coot is much less secretive than most of the rail family, and can be seen swimming on open water or\\ 
            &  walking across waterside grasslands.\\
        0.588 &  In Europe, there are colonies all along the Mediterranean coast, and also on the Atlantic islands and coasts\\ 
            &  north to Brittany and west to the Azores.\\
        0.400 &  The goosander is one of the species to which the Agreement on the Conservation of African-Eurasian Migratory \\ 
            & Waterbirds applies.\\
        0.318 &  These dense concentrations of birds are thought to be a defense against attacks by birds of prey such as \\ 
            & peregrine falcons or Eurasian sparrowhawks.\\
        \\ 
    \end{tabular}
\end{minipage}
    \caption{Comparison of top-5 sentences scores given by the pretrained and fine-tuned SkyCLIP model on  samples of the WikiRS dataset. Scores are cosine similarity values between the text and images features. Scores are ranked by decreasing order of magnitude.}
    \label{fig:simi_suppl}
\end{figure*}

%% file: table9_dataset_samples.tex
\begin{table*}[ht]
\footnotesize
    \centering
    \begin{tabular}{cc}
        \begin{minipage}{.15\textwidth}
                \includegraphics[width=1\linewidth, ]{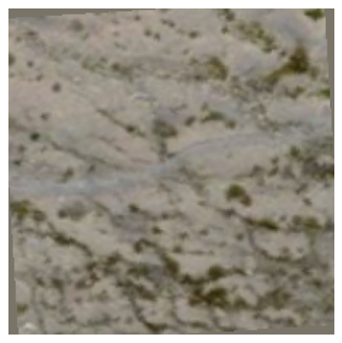} 
                
        \end{minipage}
    &   \begin{minipage}{0.8\textwidth}
                \textbf{Observed species : Saxifraga aizoides}
               \begin{itemize}
                   \item It prefers cold and moist well-draining neutral to basic bedrock, gravel, sand, or shale cliff environments.
                   \item  It is found in North America, including Alaska, across Canada, the Great Lakes region, and Greenland, and in Europe, including the Tatra Mountains, Alps, and Svalbard. 
                   \item It is a listed threatened species in New York state. 
               \end{itemize}
        \end{minipage}
    \\ 
        \begin{minipage}{.15\textwidth}
                \includegraphics[width=1\linewidth, ]{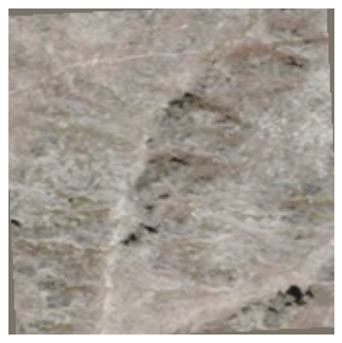} 
        \end{minipage}
    &   \begin{minipage}{0.8\textwidth}
                \textbf{Observed species : Capra ibex}
               \begin{itemize}
                   \item Alpine ibex are now found in most or all of the Italian and French alpine ranges, southern Germany, Switzerland and Austria. 
                   \item  Alpine ibex are typically absent from woodland areas, Males use lowland meadows during the spring, which is when snow melts and green grass appears. 
                   \item  Alpine ibex tend to live in steep, rough terrain near the snow line.     
                   \item ...
               \end{itemize}
        \end{minipage} 
    \\ 
        \begin{minipage}{.15\textwidth}
                \includegraphics[width=\linewidth, ]{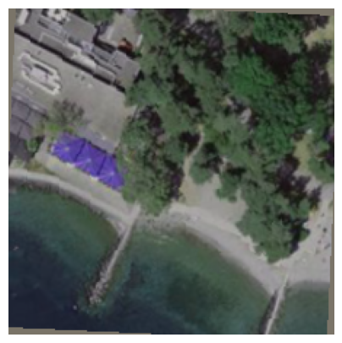} 
        \end{minipage}
    &   \begin{minipage}{0.8\textwidth}
               \textbf{ Observed species : Sedum acre}
               \begin{itemize}
                   \item It is specially adapted for growing on thin dry soils and can be found on shingle, beaches, drystone walls, dry banks, seashore rocks, roadside verges, wasteland and in sandy meadows near the sea.  
                   \item  Biting stonecrop is a tufted evergreen perennial that forms mat-like stands some tall.
                   \item Biting stonecrop spreads when allowed to do so, but is easily controlled, being shallow-rooted.
                   \item It is used in hanging baskets and container gardens, as a trailing accent, in borders, or as groundcover.
                   \item This plant grows as a creeping ground cover, often in dry sandy soil, but also in the cracks of masonry. 
                   \item  It grows well in poor soils, sand, rock gardens, and rich garden soil, under a variety of light levels.
                   \item Biting stonecrop is said to have a peppery taste (hence the name "biting") and is sometimes used in herbal medicine.
                   \item  However, it is considered to be poisonous and consumption is discouraged.                   
               \end{itemize}
        \end{minipage}
    \\
            \begin{minipage}{.15\textwidth}
                \includegraphics[width=\linewidth, ]{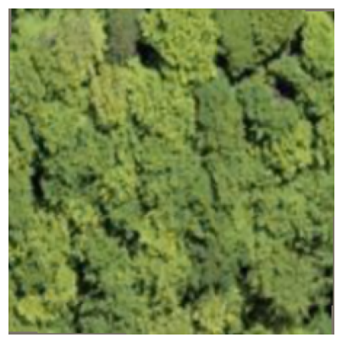} 
        \end{minipage}
    &   \begin{minipage}{0.8\textwidth}
               \textbf{ Observed species :  Cardamine heptaphylla }
               \begin{itemize}
               \item This species is widespread in Central and Southern Europe, from Northern Spain, to Italy and S.W. Germany.
                \item This species grows mainly in mountain woods, especially in beech and spruce forests, but sometimes in plain, at an elevation up to 2,200 metres (7,200 ft) above sea level. 
                \item It prefers calcareous soils. 
               \end{itemize}
        \end{minipage}
    \\
        \begin{minipage}{.15\textwidth}
                \includegraphics[width=\linewidth, ]{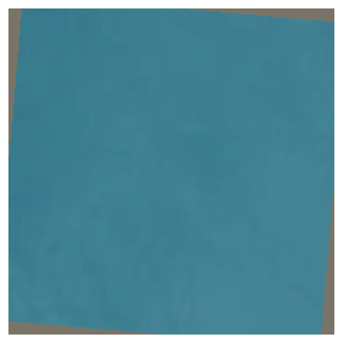} 
        \end{minipage}
    &   \begin{minipage}{0.8\textwidth}
                \textbf{Observed species : Fulica atra}
                \begin{itemize}
                    \item The coot breeds across much of the Old World on freshwater lakes and ponds, and like its relative the common moorhen, has adapted well to living in urban environments, often being found in parks and gardens with access to water.
                    \item It occurs and breeds in Europe, Asia, Australia, and Africa.
                    \item The Eurasian coot is much less secretive than most of the rail family, and can be seen swimming on open water or walking across waterside grasslands. 
                    \item It is an aggressive species, and strongly territorial during the breeding season, and both parents are involved in territorial defence.
                    \item They form large flocks on open water in winter. 
                    \item They are also found on coastal lagoons, shorelines and sheltered ponds.
                    \item ...
                \end{itemize}
        \end{minipage}
        \\
        \begin{minipage}{.15\textwidth}
                \includegraphics[width=\linewidth, ]{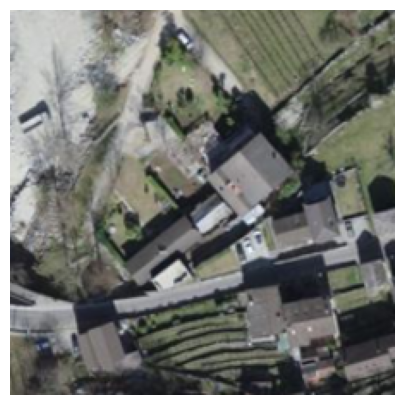} 
        \end{minipage}
    &   \begin{minipage}{0.8\textwidth}
                \textbf{Observed species: Fringilla coelebs, Sambucus nigra, Turdus merula}
                \begin{itemize}
                    \item The common chaffinch breeds in wooded areas where the July isotherm is between between 12 and 30 C.
                    \item The breeding range includes northwestern Africa and most of Europe and extends eastwards across temperate Asia to the Angara River and the southern end of Lake Baikal in Siberia.
                    \item Hedges, waste-ground roadsides, and woods are the typical habitats for the species.
                    \item S. nigra is recorded as very common in Ireland in hedges as scrub in woods.
                    \item ...
                \end{itemize}
        \end{minipage}
    \end{tabular}%
    \caption{Image-text pairs taken taken from our dataset with habitat sentences.}
    \label{tab:dataset_samples}
\end{table*}

\clearpage

%% file: main.bbl
\begin{thebibliography}{61}
\providecommand{\natexlab}[1]{#1}
\providecommand{\url}[1]{\texttt{#1}}
\expandafter\ifx\csname urlstyle\endcsname\relax
  \providecommand{\doi}[1]{doi: #1}\else
  \providecommand{\doi}{doi: \begingroup \urlstyle{rm}\Url}\fi

\bibitem[iNa()]{iNaturalist}
{iNaturalist}.
\newblock \url{https://www.inaturalist.org}.
\newblock Accessed: 2025-03-01.

\bibitem[Allingham et~al.(2023)Allingham, Ren, Dusenberry, Gu, Cui, Tran, Liu, and Lakshminarayanan]{allingham2023simple}
James~Urquhart Allingham, Jie Ren, Michael~W Dusenberry, Xiuye Gu, Yin Cui, Dustin Tran, Jeremiah~Zhe Liu, and Balaji Lakshminarayanan.
\newblock A simple zero-shot prompt weighting technique to improve prompt ensembling in text-image models.
\newblock In \emph{International Conference on Machine Learning}, pages 547--568. PMLR, 2023.

\bibitem[Berg et~al.(2014)Berg, Liu, Woo~Lee, Alexander, Jacobs, and Belhumeur]{berg2014birdsnap}
Thomas Berg, Jiongxin Liu, Seung Woo~Lee, Michelle~L Alexander, David~W Jacobs, and Peter~N Belhumeur.
\newblock Birdsnap: Large-scale fine-grained visual categorization of birds.
\newblock In \emph{Proceedings of the IEEE conference on computer vision and pattern recognition}, pages 2011--2018, 2014.

\bibitem[Brandt et~al.(2020)Brandt, Tucker, Kariryaa, Rasmussen, Abel, Small, Chave, Rasmussen, Hiernaux, Diouf, et~al.]{brandt2020unexpectedly}
Martin Brandt, Compton~J Tucker, Ankit Kariryaa, Kjeld Rasmussen, Christin Abel, Jennifer Small, Jerome Chave, Laura~Vang Rasmussen, Pierre Hiernaux, Abdoul~Aziz Diouf, et~al.
\newblock An unexpectedly large count of trees in the west african sahara and sahel.
\newblock \emph{Nature}, 587\penalty0 (7832):\penalty0 78--82, 2020.

\bibitem[Callaghan et~al.(2023)Callaghan, Borda-de {\'A}gua, van Klink, Rozzi, and Pereira]{callaghan2023unveiling}
Corey~T Callaghan, Lu{\'\i}s Borda-de {\'A}gua, Roel van Klink, Roberto Rozzi, and Henrique~M Pereira.
\newblock Unveiling global species abundance distributions.
\newblock \emph{Nature ecology \& evolution}, 7\penalty0 (10):\penalty0 1600--1609, 2023.

\bibitem[Chappuis et~al.(2022)Chappuis, Zermatten, Lobry, Le~Saux, and Tuia]{chappuis2022prompt}
Christel Chappuis, Val{\'e}rie Zermatten, Sylvain Lobry, Bertrand Le~Saux, and Devis Tuia.
\newblock Prompt-rsvqa: Prompting visual context to a language model for remote sensing visual question answering.
\newblock In \emph{Proceedings of the IEEE/CVF Conference on Computer Vision and Pattern Recognition}, pages 1372--1381, 2022.

\bibitem[Cheng et~al.(2021)Cheng, Wu, Zhang, Vajda, and Gonzalez]{cheng2021data}
Ruizhe Cheng, Bichen Wu, Peizhao Zhang, Peter Vajda, and Joseph~E Gonzalez.
\newblock Data-efficient language-supervised zero-shot learning with self-distillation.
\newblock In \emph{Proceedings of the IEEE/CVF Conference on Computer Vision and Pattern Recognition}, pages 3119--3124, 2021.

\bibitem[Chu et~al.(2024)Chu, Zheng, Ji, Wang, and Chua]{chu_towards_2024_geotext}
Meng Chu, Zhedong Zheng, Wei Ji, Tingyu Wang, and Tat-Seng Chua.
\newblock Towards natural language-guided drones: Geotext-1652 benchmark with spatial relation matching.
\newblock In \emph{European Conference on Computer Vision}, pages 213--231. Springer, 2024.

\bibitem[Chytr{\`y} et~al.(2020)Chytr{\`y}, Tich{\`y}, Hennekens, Knollov{\'a}, Janssen, Rodwell, Peterka, Marcen{\`o}, Landucci, Danihelka, et~al.]{chytry2020eunis}
Milan Chytr{\`y}, Lubom{\'\i}r Tich{\`y}, Stephan~M Hennekens, Ilona Knollov{\'a}, John~AM Janssen, John~S Rodwell, Tom{\'a}{\v{s}} Peterka, Corrado Marcen{\`o}, Flavia Landucci, Ji{\v{r}}{\'\i} Danihelka, et~al.
\newblock Eunis habitat classification: Expert system, characteristic species combinations and distribution maps of european habitats.
\newblock \emph{Applied Vegetation Science}, 23\penalty0 (4):\penalty0 648--675, 2020.

\bibitem[Daroya et~al.(2024)Daroya, Cole, Aodha, Horn, and Maji]{daroya_wildsat_2024}
Rangel Daroya, Elijah Cole, Oisin~Mac Aodha, Grant~Van Horn, and Subhransu Maji.
\newblock {WildSAT}: {Learning} {Satellite} {Image} {Representations} from {Wildlife} {Observations}, 2024.
\newblock arXiv:2412.14428 [cs].

\bibitem[Daru et~al.(2018)Daru, Park, Primack, Willis, Barrington, Whitfeld, Seidler, Sweeney, Foster, Ellison, et~al.]{daru2018widespread}
Barnabas~H Daru, Daniel~S Park, Richard~B Primack, Charles~G Willis, David~S Barrington, Timothy~JS Whitfeld, Tristram~G Seidler, Patrick~W Sweeney, David~R Foster, Aaron~M Ellison, et~al.
\newblock Widespread sampling biases in herbaria revealed from large-scale digitization.
\newblock \emph{New Phytologist}, 217\penalty0 (2):\penalty0 939--955, 2018.

\bibitem[Delplanque et~al.(2022)Delplanque, Foucher, Lejeune, Linchant, and Th{\'e}au]{delplanque2022multispecies}
Alexandre Delplanque, Samuel Foucher, Philippe Lejeune, Julie Linchant, and J{\'e}r{\^o}me Th{\'e}au.
\newblock Multispecies detection and identification of african mammals in aerial imagery using convolutional neural networks.
\newblock \emph{Remote Sensing in Ecology and Conservation}, 8\penalty0 (2):\penalty0 166--179, 2022.

\bibitem[Dhakal et~al.(2024)Dhakal, Ahmad, Khanal, Sastry, Kerner, and Jacobs]{dhakal2024sat2cap}
Aayush Dhakal, Adeel Ahmad, Subash Khanal, Srikumar Sastry, Hannah Kerner, and Nathan Jacobs.
\newblock Sat2cap: Mapping fine-grained textual descriptions from satellite images.
\newblock In \emph{Proceedings of the IEEE/CVF Conference on Computer Vision and Pattern Recognition}, pages 533--542, 2024.

\bibitem[Dwibedi et~al.(2021)Dwibedi, Aytar, Tompson, Sermanet, and Zisserman]{dwibedi_little_2021}
Debidatta Dwibedi, Yusuf Aytar, Jonathan Tompson, Pierre Sermanet, and Andrew Zisserman.
\newblock With a {Little} {Help} from {My} {Friends}: {Nearest}-{Neighbor} {Contrastive} {Learning} of {Visual} {Representations}.
\newblock In \emph{2021 {IEEE}/{CVF} {International} {Conference} on {Computer} {Vision} ({ICCV})}, Montreal, QC, Canada, 2021. IEEE.

\bibitem[El~Banani et~al.(2023)El~Banani, Desai, and Johnson]{el2023language_guided}
Mohamed El~Banani, Karan Desai, and Justin Johnson.
\newblock Learning visual representations via language-guided sampling.
\newblock In \emph{Proceedings of the IEEE/CVF Conference on Computer Vision and Pattern Recognition}, pages 19208--19220, 2023.

\bibitem[Fithian et~al.(2015)Fithian, Elith, Hastie, and Keith]{fithian2015bias}
William Fithian, Jane Elith, Trevor Hastie, and David~A Keith.
\newblock Bias correction in species distribution models: pooling survey and collection data for multiple species.
\newblock \emph{Methods in ecology and evolution}, 6\penalty0 (4):\penalty0 424--438, 2015.

\bibitem[Foundation(2025)]{wikipedia2025}
Wikimedia Foundation.
\newblock Wikipedia corpus.
\newblock Wikimedia Foundation, The Free Encyclopedia, 2025.
\newblock Accessed from \url{https://dumps.wikimedia.org/}.

\bibitem[Gao et~al.(2024)Gao, Liu, Xu, Wu, Zhang, Li, Yang, Liu, and Sun]{gao_softclip_2024}
Yuting Gao, Jinfeng Liu, Zihan Xu, Tong Wu, Enwei Zhang, Ke Li, Jie Yang, Wei Liu, and Xing Sun.
\newblock {SoftCLIP}: {Softer} {Cross}-{Modal} {Alignment} {Makes} {CLIP} {Stronger}.
\newblock \emph{Proceedings of the AAAI Conference on Artificial Intelligence}, 38\penalty0 (3):\penalty0 1860--1868, 2024.
\newblock Number: 3.

\bibitem[{GBIF.Org User}(2024)]{gbif_download}
{GBIF.Org User}.
\newblock Gbif occurrence download, doi:10.15468/dl.pr8cws, 2024.

\bibitem[Go{\"e}au et~al.(2013)Go{\"e}au, Bonnet, Joly, Baki{\'c}, Barbe, Yahiaoui, Selmi, Carr{\'e}, Barth{\'e}l{\'e}my, Boujemaa, et~al.]{goeau2013plantnet}
Herv{\'e} Go{\"e}au, Pierre Bonnet, Alexis Joly, Vera Baki{\'c}, Julien Barbe, Itheri Yahiaoui, Souheil Selmi, Jennifer Carr{\'e}, Daniel Barth{\'e}l{\'e}my, Nozha Boujemaa, et~al.
\newblock Pl@ ntnet mobile app.
\newblock In \emph{Proceedings of the 21st ACM international conference on Multimedia}, pages 423--424, 2013.

\bibitem[Hamilton et~al.(2024)Hamilton, Lange, Cole, Shepard, Heinrich, Mac~Aodha, Van~Horn, and Maji]{hamilton_2024_combining}
Max Hamilton, Christian Lange, Elijah Cole, Alexander Shepard, Samuel Heinrich, Oisin Mac~Aodha, Grant Van~Horn, and Subhransu Maji.
\newblock Combining observational data and language for species range estimation.
\newblock \emph{Advances in Neural Information Processing Systems}, 37:\penalty0 17719--17742, 2024.

\bibitem[Ilharco et~al.(2021)Ilharco, Wortsman, Wightman, Gordon, Carlini, Taori, Dave, Shankar, Namkoong, Miller, Hajishirzi, Farhadi, and Schmidt]{ilharco_gabriel_openclip}
Gabriel Ilharco, Mitchell Wortsman, Ross Wightman, Cade Gordon, Nicholas Carlini, Rohan Taori, Achal Dave, Vaishaal Shankar, Hongseok Namkoong, John Miller, Hannaneh Hajishirzi, Ali Farhadi, and Ludwig Schmidt.
\newblock Openclip, 2021.

\bibitem[Jain et~al.(2025)Jain, Ienco, Interdonato, Berchoux, and Marcos]{jain_senclip_2024}
Pallavi Jain, Dino Ienco, Roberto Interdonato, Tristan Berchoux, and Diego Marcos.
\newblock Senclip: Enhancing zero-shot land-use mapping for sentinel-2 with ground-level prompting.
\newblock In \emph{Proceedings of the Winter Conference on Applications of Computer Vision (WACV)}, pages 5656--5665, 2025.

\bibitem[Jia et~al.(2021)Jia, Yang, Xia, Chen, Parekh, Pham, Le, Sung, Li, and Duerig]{jia2021ALIGN}
Chao Jia, Yinfei Yang, Ye Xia, Yi-Ting Chen, Zarana Parekh, Hieu Pham, Quoc Le, Yun-Hsuan Sung, Zhen Li, and Tom Duerig.
\newblock Scaling up visual and vision-language representation learning with noisy text supervision.
\newblock In \emph{International conference on machine learning}, pages 4904--4916. PMLR, 2021.

\bibitem[Kirchhoff et~al.(2021)Kirchhoff, Callaghan, Keith, Indiarto, Taseski, Ooi, Le~Breton, Mesaglio, Kingsford, and Cornwell]{kirchhoff2021rapidly}
Casey Kirchhoff, Corey~T Callaghan, David~A Keith, Dony Indiarto, Guy Taseski, Mark~KJ Ooi, Tom~D Le~Breton, Thomas Mesaglio, Richard~T Kingsford, and William~K Cornwell.
\newblock Rapidly mapping fire effects on biodiversity at a large-scale using citizen science.
\newblock \emph{Science of the Total environment}, 755:\penalty0 142348, 2021.

\bibitem[Lang et~al.(2023)Lang, Jetz, Schindler, and Wegner]{lang2023high}
Nico Lang, Walter Jetz, Konrad Schindler, and Jan~Dirk Wegner.
\newblock A high-resolution canopy height model of the earth.
\newblock \emph{Nature Ecology \& Evolution}, 7\penalty0 (11):\penalty0 1778--1789, 2023.

\bibitem[Li et~al.(2022{\natexlab{a}})Li, Li, Xiong, and Hoi]{li2022blip}
Junnan Li, Dongxu Li, Caiming Xiong, and Steven Hoi.
\newblock Blip: Bootstrapping language-image pre-training for unified vision-language understanding and generation.
\newblock In \emph{International conference on machine learning}, pages 12888--12900. PMLR, 2022{\natexlab{a}}.

\bibitem[Li et~al.(2023)Li, Li, Savarese, and Hoi]{li2023blip2}
Junnan Li, Dongxu Li, Silvio Savarese, and Steven Hoi.
\newblock Blip-2: Bootstrapping language-image pre-training with frozen image encoders and large language models.
\newblock In \emph{International conference on machine learning}, pages 19730--19742. PMLR, 2023.

\bibitem[Li et~al.(2022{\natexlab{b}})Li, Xia, Ge, and Liu]{li2022selective}
Shikun Li, Xiaobo Xia, Shiming Ge, and Tongliang Liu.
\newblock Selective-supervised contrastive learning with noisy labels.
\newblock In \emph{Proceedings of the IEEE/CVF conference on computer vision and pattern recognition}, pages 316--325, 2022{\natexlab{b}}.

\bibitem[Li et~al.(2022{\natexlab{c}})Li, Liang, Zhao, Cui, Ouyang, Shao, Yu, and Yan]{li_supervision_2022_declip}
Yangguang Li, Feng Liang, Lichen Zhao, Yufeng Cui, Wanli Ouyang, Jing Shao, Fengwei Yu, and Junjie Yan.
\newblock Supervision {Exists} {Everywhere}: {A} {Data} {Efficient} {Contrastive} {Language}-{Image} {Pre}-training {Paradigm}.
\newblock In \emph{International Conference on Learning Representations}, 2022{\natexlab{c}}.

\bibitem[Liu et~al.(2024)Liu, Chen, Guan, Zhou, Zhu, Ye, Fu, and Zhou]{liu2024remoteclip}
Fan Liu, Delong Chen, Zhangqingyun Guan, Xiaocong Zhou, Jiale Zhu, Qiaolin Ye, Liyong Fu, and Jun Zhou.
\newblock Remoteclip: A vision language foundation model for remote sensing.
\newblock \emph{IEEE Transactions on Geoscience and Remote Sensing}, 2024.

\bibitem[Loshchilov and Hutter(2017)]{loshchilov2017ADAMWdecoupled}
Ilya Loshchilov and Frank Hutter.
\newblock Decoupled weight decay regularization.
\newblock In \emph{International Conference on Learning Representations}, 2017.

\bibitem[Mac~Aodha et~al.(2019)Mac~Aodha, Cole, and Perona]{mac2019presence}
Oisin Mac~Aodha, Elijah Cole, and Pietro Perona.
\newblock Presence-only geographical priors for fine-grained image classification.
\newblock In \emph{Proceedings of the IEEE/CVF International Conference on Computer Vision}, pages 9596--9606, 2019.

\bibitem[Morgado et~al.(2021)Morgado, Misra, and Vasconcelos]{morgado2021robust}
Pedro Morgado, Ishan Misra, and Nuno Vasconcelos.
\newblock Robust audio-visual instance discrimination.
\newblock In \emph{Proceedings of the IEEE/CVF Conference on Computer Vision and Pattern Recognition}, pages 12934--12945, 2021.

\bibitem[Mu et~al.(2022)Mu, Kirillov, Wagner, and Xie]{mu_slip_2021}
Norman Mu, Alexander Kirillov, David Wagner, and Saining Xie.
\newblock Slip: Self-supervision meets language-image pre-training.
\newblock In \emph{European conference on computer vision}, pages 529--544. Springer, 2022.

\bibitem[Muhtar et~al.(2024)Muhtar, Li, Gu, Zhang, and Xiao]{muhtar2024lhrs}
Dilxat Muhtar, Zhenshi Li, Feng Gu, Xueliang Zhang, and Pengfeng Xiao.
\newblock Lhrs-bot: Empowering remote sensing with vgi-enhanced large multimodal language model.
\newblock In \emph{European Conference on Computer Vision}, pages 440--457. Springer, 2024.

\bibitem[Oord et~al.(2018)Oord, Li, and Vinyals]{oord2018representation}
Aaron van~den Oord, Yazhe Li, and Oriol Vinyals.
\newblock Representation learning with contrastive predictive coding.
\newblock \emph{arXiv preprint arXiv:1807.03748}, 2018.

\bibitem[Paz-Argaman et~al.(2020)Paz-Argaman, Tsarfaty, Chechik, and Atzmon]{paz-argaman_zest_2020}
Tzuf Paz-Argaman, Reut Tsarfaty, Gal Chechik, and Yuval Atzmon.
\newblock {ZEST}: {Zero}-shot {Learning} from {Text} {Descriptions} using {Textual} {Similarity} and {Visual} {Summarization}.
\newblock In \emph{Findings of the {Association} for {Computational} {Linguistics}: {EMNLP} 2020}, pages 569--579, Online, 2020. Association for Computational Linguistics.

\bibitem[Radford et~al.(2021)Radford, Kim, Hallacy, Ramesh, Goh, Agarwal, Sastry, Askell, Mishkin, Clark, et~al.]{radford2021learning}
Alec Radford, Jong~Wook Kim, Chris Hallacy, Aditya Ramesh, Gabriel Goh, Sandhini Agarwal, Girish Sastry, Amanda Askell, Pamela Mishkin, Jack Clark, et~al.
\newblock Learning transferable visual models from natural language supervision.
\newblock In \emph{International conference on machine learning}, pages 8748--8763. PMLR, 2021.

\bibitem[Reed et~al.(2014)Reed, Lee, Anguelov, Szegedy, Erhan, and Rabinovich]{reed2014training}
Scott Reed, Honglak Lee, Dragomir Anguelov, Christian Szegedy, Dumitru Erhan, and Andrew Rabinovich.
\newblock Training deep neural networks on noisy labels with bootstrapping.
\newblock \emph{arXiv preprint arXiv:1412.6596}, 2014.

\bibitem[Ricci et~al.(2024)Ricci, Bazi, and Melgani]{ricci2024machine}
Riccardo Ricci, Yakoub Bazi, and Farid Melgani.
\newblock Machine-to-machine visual dialoguing with chatgpt for enriched textual image description.
\newblock \emph{Remote Sensing}, 16\penalty0 (3):\penalty0 441, 2024.

\bibitem[Richardson(2007)]{richardson2007beautiful}
Leonard Richardson.
\newblock Beautiful soup documentation.
\newblock \emph{April}, 2007.

\bibitem[Robertson et~al.(2014)Robertson, D{\"o}ring, Guralnick, Bloom, Wieczorek, Braak, Otegui, Russell, and Desmet]{robertson2014gbif}
Tim Robertson, Markus D{\"o}ring, Robert Guralnick, David Bloom, John Wieczorek, Kyle Braak, Javier Otegui, Laura Russell, and Peter Desmet.
\newblock The gbif integrated publishing toolkit: facilitating the efficient publishing of biodiversity data on the internet.
\newblock \emph{PloS one}, 9\penalty0 (8):\penalty0 e102623, 2014.

\bibitem[Saha et~al.(2024)Saha, Van~Horn, and Maji]{saha_improved_2024}
Oindrila Saha, Grant Van~Horn, and Subhransu Maji.
\newblock Improved {Zero}-{Shot} {Classification} by {Adapting} {VLMs} with {Text} {Descriptions}.
\newblock In \emph{2024 {IEEE}/{CVF} {Conference} on {Computer} {Vision} and {Pattern} {Recognition} ({CVPR})}, pages 17542--17552, Seattle, WA, USA, 2024. IEEE.

\bibitem[Sastry et~al.(2024)Sastry, Khanal, Dhakal, Ahmad, and Jacobs]{sastry_taxabind_2024}
Srikumar Sastry, Subash Khanal, Aayush Dhakal, Adeel Ahmad, and Nathan Jacobs.
\newblock {TaxaBind}: {A} {Unified} {Embedding} {Space} for {Ecological} {Applications}, 2024.
\newblock arXiv:2411.00683 [cs].

\bibitem[Schuhmann et~al.(2022)Schuhmann, Beaumont, Vencu, Gordon, Wightman, Cherti, Coombes, Katta, Mullis, Wortsman, et~al.]{schuhmann2022laion}
Christoph Schuhmann, Romain Beaumont, Richard Vencu, Cade Gordon, Ross Wightman, Mehdi Cherti, Theo Coombes, Aarush Katta, Clayton Mullis, Mitchell Wortsman, et~al.
\newblock Laion-5b: An open large-scale dataset for training next generation image-text models.
\newblock \emph{Advances in neural information processing systems}, 35:\penalty0 25278--25294, 2022.

\bibitem[Shabbir et~al.(2025)Shabbir, Zumri, Bennamoun, Khan, and Khan]{shabbir2025geopixel}
Akashah Shabbir, Mohammed Zumri, Mohammed Bennamoun, Fahad~S Khan, and Salman Khan.
\newblock Geopixel: Pixel grounding large multimodal model in remote sensing.
\newblock \emph{arXiv preprint arXiv:2501.13925}, 2025.

\bibitem[Sullivan et~al.(2014)Sullivan, Aycrigg, Barry, Bonney, Bruns, Cooper, Damoulas, Dhondt, Dietterich, Farnsworth, et~al.]{sullivan2014ebird}
Brian~L Sullivan, Jocelyn~L Aycrigg, Jessie~H Barry, Rick~E Bonney, Nicholas Bruns, Caren~B Cooper, Theo Damoulas, Andr{\'e}~A Dhondt, Tom Dietterich, Andrew Farnsworth, et~al.
\newblock The ebird enterprise: An integrated approach to development and application of citizen science.
\newblock \emph{Biological conservation}, 169:\penalty0 31--40, 2014.

\bibitem[Uzkent et~al.(2019)Uzkent, Sheehan, Meng, Tang, Burke, Lobell, and Ermon]{uzkent_learning_2019}
Burak Uzkent, Evan Sheehan, Chenlin Meng, Zhongyi Tang, Marshall Burke, David Lobell, and Stefano Ermon.
\newblock Learning to {Interpret} {Satellite} {Images} using {Wikipedia}.
\newblock \emph{Proceedings of the Twenty-Eighth International Joint Conference on Artificial Intelligence}, 2019.

\bibitem[van Tiel et~al.(2024)van Tiel, Fopp, Brun, van~den Hoogen, Karger, Casadei, Lyu, Tuia, Zimmermann, Crowther, et~al.]{van2024regional}
Nina van Tiel, Fabian Fopp, Philipp Brun, Johan van~den Hoogen, Dirk~Nikolaus Karger, Cecilia~M Casadei, Lisha Lyu, Devis Tuia, Niklaus~E Zimmermann, Thomas~W Crowther, et~al.
\newblock Regional uniqueness of tree species composition and response to forest loss and climate change.
\newblock \emph{Nature Communications}, 15\penalty0 (1):\penalty0 4375, 2024.

\bibitem[Vardi et~al.(2021)Vardi, Berger-Tal, and Roll]{vardi2021inaturalist}
Reut Vardi, Oded Berger-Tal, and Uri Roll.
\newblock inaturalist insights illuminate covid-19 effects on large mammals in urban centers.
\newblock \emph{Biological conservation}, 254:\penalty0 108953, 2021.

\bibitem[Wang et~al.(2024)Wang, Prabha, Huang, Wu, and Rajagopal]{wang_skyscript_2024}
Zhecheng Wang, Rajanie Prabha, Tianyuan Huang, Jiajun Wu, and Ram Rajagopal.
\newblock {SkyScript}: {A} {Large} and {Semantically} {Diverse} {Vision}-{Language} {Dataset} for {Remote} {Sensing}.
\newblock \emph{Proceedings of the AAAI Conference on Artificial Intelligence}, 38\penalty0 (6):\penalty0 5805--5813, 2024.
\newblock Number: 6.

\bibitem[Wei et~al.(2023)Wei, Jiang, Huang, and Sun]{wei2023instructiongpt}
Lai Wei, Zihao Jiang, Weiran Huang, and Lichao Sun.
\newblock Instructiongpt-4: A 200-instruction paradigm for fine-tuning minigpt-4.
\newblock \emph{arXiv preprint arXiv:2308.12067}, 2023.

\bibitem[Weiss and Banko(2018)]{weiss2018ecosystem}
Michael Weiss and GJTP Banko.
\newblock Ecosystem type map v3. 1--terrestrial and marine ecosystems.
\newblock \emph{European Topic Centre on Biological Diversity report to the European Environment Agency (EEA)}, page~79, 2018.

\bibitem[Wolf et~al.(2022)Wolf, Mahecha, Sabatini, Wirth, Bruelheide, Kattge, Moreno~Mart{\'\i}nez, Mora, and Kattenborn]{wolf2022citizen}
Sophie Wolf, Miguel~D Mahecha, Francesco~Maria Sabatini, Christian Wirth, Helge Bruelheide, Jens Kattge, {\'A}lvaro Moreno~Mart{\'\i}nez, Karin Mora, and Teja Kattenborn.
\newblock Citizen science plant observations encode global trait patterns.
\newblock \emph{Nature ecology \& evolution}, 6\penalty0 (12):\penalty0 1850--1859, 2022.

\bibitem[Wu et~al.(2023)Wu, Zhang, Gu, Duporge, Hughey, Stabach, Skidmore, Hopcraft, Lee, Atkinson, et~al.]{wu2023deep}
Zijing Wu, Ce Zhang, Xiaowei Gu, Isla Duporge, Lacey~F Hughey, Jared~A Stabach, Andrew~K Skidmore, J~Grant~C Hopcraft, Stephen~J Lee, Peter~M Atkinson, et~al.
\newblock Deep learning enables satellite-based monitoring of large populations of terrestrial mammals across heterogeneous landscape.
\newblock \emph{Nature communications}, 14\penalty0 (1):\penalty0 3072, 2023.

\bibitem[Yang et~al.(2023)Yang, Deng, An, Li, Feng, Guo, Yang, and Liu]{yang_alip_2023}
Kaicheng Yang, Jiankang Deng, Xiang An, Jiawei Li, Ziyong Feng, Jia Guo, Jing Yang, and Tongliang Liu.
\newblock Alip: Adaptive language-image pre-training with synthetic caption.
\newblock In \emph{Proceedings of the IEEE/CVF International Conference on Computer Vision}, pages 2922--2931, 2023.

\bibitem[Zermatten et~al.(2025)Zermatten, Castillo-Navarro, Marcos, and Tuia]{zermatten2025learning}
Val{\'e}rie Zermatten, Javiera Castillo-Navarro, Diego Marcos, and Devis Tuia.
\newblock Learning transferable land cover semantics for open vocabulary interactions with remote sensing images.
\newblock \emph{ISPRS Journal of Photogrammetry and Remote Sensing}, 220:\penalty0 621--636, 2025.

\bibitem[Zhang et~al.(2018)Zhang, Cisse, Dauphin, and Lopez-Paz]{zhang2018mixup}
Hongyi Zhang, Moustapha Cisse, Yann~N Dauphin, and David Lopez-Paz.
\newblock mixup: Beyond empirical risk minimization.
\newblock In \emph{International Conference on Learning Representations}, 2018.

\bibitem[Zhang et~al.(2024)Zhang, Zhao, Guo, and Yin]{zhang_georsclip_2024}
Zilun Zhang, Tiancheng Zhao, Yulong Guo, and Jianwei Yin.
\newblock {RS5M} and {GeoRSCLIP}: {A} {Large} {Scale} {Vision}-{Language} {Dataset} and {A} {Large} {Vision}-{Language} {Model} for {Remote} {Sensing}.
\newblock \emph{IEEE Transactions on Geoscience and Remote Sensing}, pages 1--1, 2024.
\newblock arXiv:2306.11300 [cs].

\bibitem[Zizka et~al.(2020)Zizka, Carvalho, Calvente, Baez-Lizarazo, Cabral, Coelho, Colli-Silva, Fantinati, Fernandes, Ferreira-Ara{\'u}jo, et~al.]{zizka2020no}
Alexander Zizka, Fernanda~Antunes Carvalho, Alice Calvente, Mabel~Rocio Baez-Lizarazo, Andressa Cabral, J{\'e}ssica Fernanda~Ramos Coelho, Matheus Colli-Silva, Mariana~Ramos Fantinati, Moabe~F Fernandes, Thais Ferreira-Ara{\'u}jo, et~al.
\newblock No one-size-fits-all solution to clean gbif.
\newblock \emph{PeerJ}, 8:\penalty0 e9916, 2020.

\end{thebibliography}
